\documentclass[12pt]{article}

\newcommand{\blind}{0}

\usepackage{amsthm,amsmath,amssymb,bbm}
\usepackage{pdfpages}

\usepackage{setspace, fancybox, fullpage}
\usepackage{comment}
\usepackage{multirow}
\usepackage{rotating}
\usepackage{subfigure}
\usepackage[round]{natbib} 
\usepackage{colortbl}

\newtheorem{theorem}{Theorem}

\newtheorem{corollary}{Corollary}
\newtheorem{proposition}{Proposition}
\newtheorem{assumption}{Assumption}

\newcommand{\bK}{\boldsymbol{K}}
\newcommand{\bX}{\boldsymbol{X}}
\newcommand{\bx}{\boldsymbol{x}}

\newcommand{\bbeta}{\boldsymbol{\beta}}
\newcommand{\btheta}{\boldsymbol{\theta}}

\newcommand{\bdf}{\boldsymbol{f}}

\newcommand{\be}{\boldsymbol{e}}
\newcommand{\bone}{\boldsymbol{1}}

\newcommand{\mbbR}{\mathbb{R}}
\newcommand{\mbbP}{\mathbb{P}}
\newcommand{\mF}{\mathcal{F}}

\newcommand{\mY}{\mathcal{Y}}

\newcommand{\R}{\mathbb{R}}
\newcommand{\E}{\mathbb{E}}
\newcommand{\ind}[1]{\mathbbm{1}_{\left[ {#1} \right] }}
\newcommand{\Ind}[1]{\mathbbm{1}_{\left\{ {#1} \right\} }}

\def\argmin{\mathop{\rm argmin}}
\def\argmax{\mathop{\rm argmax}}

\def\arginf{\mathop{\rm arginf}}

\def\pr{\textrm{pr}}
\def\rej{\mbox{\textregistered}}

\begin{document}

\def\spacingset#1{\renewcommand{\baselinestretch}%
{#1}\small\normalsize} \spacingset{1}

\if0\blind
{
  \title{\bf On Reject and Refine Options in Multicategory Classification}
  \author{Chong Zhang, Wenbo Wang and Xingye Qiao\thanks{Correspondence to: Xingye Qiao (e-mail: qiao@math.binghamton.edu) at Binghamton University, State University of New York, Binghamton, New York. Chong Zhang is a Machine Learning Scientist at Seattle, Washington. Wenbo Wang is a Ph.D. candidate  and Xingye Qiao is an Associate Professor  at the Department of Mathematical Sciences at Binghamton University, State University of New York, Binghamton, New York, 13902. The authors gratefully acknowledge Professor Mu Zhu for his helpful suggestions. Qiao's research is partially supported by a collaboration grant from \textit{Simons Foundation} (award number 246649). A revised version of this paper was accepted for publication in the Journal of the American Statistical Association Theory and Methods Section.}}
  \maketitle
} \fi

\if1\blind
{
  \bigskip
  \bigskip
  \bigskip
  \begin{center}
    {\LARGE\bf On Reject and Refine Options in Multicategory Classification}
  \end{center}
  \medskip
} \fi

\bigskip
\begin{abstract}
In many real applications of statistical learning, a decision made from misclassification can be too costly to afford; in this case, a reject option, which defers the decision until further investigation is conducted, is often preferred. In recent years, there has been much development for binary classification with a reject option. Yet, little progress has been made for the multicategory case. In this article, we propose margin-based multicategory classification methods with a reject option. In addition, and more importantly, we introduce a new and unique refine option for the multicategory problem, where the class of an observation is predicted to be from a set of class labels, whose cardinality is not necessarily one.  The main advantage of both options lies in their capacity of identifying error-prone observations. Moreover, the refine option can provide more constructive information for classification by effectively ruling out implausible classes. Efficient implementations have been developed for the proposed methods. On the theoretical side, we offer a novel statistical learning theory and show a fast convergence rate of the excess $\ell$-risk of our methods with emphasis on diverging dimensionality and number of classes. The results can be further improved under a low noise assumption. A set of comprehensive simulation and real data studies has shown the usefulness of the new learning tools compared to regular multicategory classifiers. Detailed proofs of theorems and extended numerical results are included in the supplemental materials available online.
\end{abstract}

\noindent%
\vfill

\newpage
\spacingset{1.45} 

\setcounter{page}{1}
\abovedisplayskip=8pt
\belowdisplayskip=8pt

\section{Introduction}
\label{introduction}

Classification is one of the founding pillars for statistical learning. In binary classification, an $i.i.d.$ training data set $\{(\bx_i,y_i),~i=1,\ldots,n\}$ is obtained from an unknown distribution $\mbbP(\bx,y)$, where $\bx\in\R^p$ is the observed covariates and $y \in \{+1,-1\}$ is the class label. The learning goal is to obtain a classifier $\phi(\cdot)$ based on the training data, such that for any new observation with only $\bx$ available, its class label can be accurately predicted using $\phi(\bx)$. The goodness of a classifier is commonly measured by the misclassification rate, $\pr \{ \phi(\bX) \neq Y \}$, where the probability is taken with respect to $\mbbP$. We aim to find the best classifier $\phi$ that minimizes the expected value of the $0$-$1$ loss $L(\bx, y, \phi)=\Ind{\phi(\bx)\neq y}$.

There are many classification methods in the literature. For an overall introduction, see \cite{Hastie2009}. Among these methods, margin-based classifiers are very popular. For a binary margin-based classifier, one typically finds a classification function $f: \mbbR^p \rightarrow \mbbR$ and defines the classifier as $\phi(\bx) = \textrm{sign} \{f(\bx)\}$. A correct classification occurs when the \textit{functional margin} $yf(\bx)>0$. Since directly minimizing the empirical $0$-$1$ loss is difficult due to the discontinuity of the $0$-$1$ loss function, a surrogate loss is often used to encourage large values of the functional margin $yf(\bx)$. Many binary margin-based classifiers using different surrogate loss functions have been proposed in the literature, such as Support Vector Machines \citep[SVM;][]{Cortes95,Vapnik98}, AdaBoost \citep{Freund97}, $\psi$-learning \citep{Shen03}, Distance-Weighted Discrimination \citep[DWD;][]{Marron07}, Large-margin Unified Machine \citep[LUM;][]{LUM} and Flexible High-dimensional Classification Machines \citep[FLAME;][]{qiao2015flexible}.

When there are $k>2$ classes, the class label $y$ can be coded as $y \in \{1,\ldots,k\}$ instead. In this article, we focus on multicategory classifiers that consider all classes simultaneously in a single optimization problem. A common approach is to train a vector-valued function $\bdf = (f_1,\ldots,f_k)^T:\R^p\mapsto\R^k$, and define the classifier as $\phi(\bx) = \argmax_{j \in \{1,\ldots,k\}} f_j(\bx)$. A sum-to-zero constraint, $\sum_{j=1}^k f_j\equiv 0$, is often imposed for theoretical and practical concerns. See, for example, \citet{Vapnik98}, \citet{Crammer01}, \citet{Lee04}, \citet{Zhu05}, \citet{multipsi06}, \citet{RMSVM}, \citet{MLUM}, among others. Recently, \citet{MAC} proposed the angle-based classification framework. The angle-based classifiers are free of the sum-to-zero constraint, and can be advantageous in terms of computational speed and classification performance, especially for high-dimensional problems. In this paper, our proposed method is based on the angle-based classification framework.

In real applications, it is often the case that an accurate decision is hard to reach, and the consequence of misclassification is disastrous and too severe to bear. In these situations, it may be wise to resort for a reject option, \textit{i.e.}, to report ``I don't know'' (denoted as $\rej$ hereafter), to avoid such a consequence. With a reject option, future resources will be allocated to these previously rejected subjects to improve their classification. For example, in cancer diagnosis, an oncologist should send a patient, who is difficult to be diagnosed based on preliminary results, for more tests, or seek a second opinion, instead of telling the patient, with little confidence, that she probably has or does not have the cancer.

To adopt a reject option, a possible approach is to modify the $0$-$1$ loss  such that when $\rej$ occurs, a positive cost is present (otherwise, $\rej$ would always be preferred). For instance, \cite{herbei2006classification} considered the $0$-$d$-$1$ loss, $L(\bx, y, \phi)=d\cdot \ind{\phi(\bx)=\rej}+\ind{\phi(\bx) \neq y,\phi(\bx)\neq\rej}$, where $d>0$ is the cost for a rejection (\textit{e.g.}, this may be the cost for the additional tests that the oncologist orders for the patient.)

Recently, there have been a number of works on the reject option for binary classification in the literature \citep{fumera2002support,herbei2006classification,wegkamp2007lasso,el2010foundations,yuan2010classification,wegkamp2011support}. However, much less attention has been paid to multicategory classification. In the literature, \citet{fumera2000reject}, \citet{tax2008growing} and \citet{le2010optimum} considered the reject option in multicategory classification using methods that depend on explicit class conditional probability estimation. However, probability or density estimation is often much more difficult than class label prediction \citep{Fuernkranz2010Preference}, especially when the dimension is high \citep{refit}. Hence, it is desirable to have a multicategory classifier with a reject option that does not rely on explicit class probability estimation. The current article fills the gap on this end.

Our first contribution is to propose multicategory classifiers with a reject option. Our methods are based on angle-based multicategory methods and do not involve estimating the class conditional probability, hence can be robust and efficient for high-dimensional problems.

Secondly, we introduce a new notion that is quite unique for the multicategory problem (which is absent in the binary case), namely, a refine option. A refinement predicts the class label to be from a set of $r$ labels, where $1\le r\le k$. When $r=1$, it reduces to the regular definite classification; when $r=k$, no information is provided and a refinement is the same as $\rej$; when $1< r< k$, we have refined the number of classes that an observation most likely belongs to, from $k$ to $r$. A smaller $r$ leads to more useful information, yet it increases the chance of misclassification. In this paper, we introduce a data-adaptive approach that can automatically select the size $r$ for a new prediction.

The usefulness of the refine option can be understood from two sides. In contrast to a definite but potentially reckless answer ($r=1$), a refinement is more cautious and risk-avert; catastrophic consequences of misclassification can be effectively avoided. On the other hand, compared with a complete reject option ($r=k$), which tells little about an observation, a refinement provides constructive information; future investigation can be conducted on a set of originally confusable classes, which can improve the classification performance.

Our next contribution is a thorough investigation of the theoretical properties of our methods, focusing on the asymptotic behavior of the excess $\ell$-risk when the number of classes $k$ and the dimension $p$ both diverge. In particular, we calibrate the difficulty of classification when $k$ increases. This helps to shed some light on the usefulness of our new refine option, that is, one can focus on a subset of classes in a refined further analysis, which can in turn improve the classification accuracy. Moreover, we demonstrate that if the number of noise predictors diverges faster than $k$ does, then the $L_1$ penalty can perform better than the $L_2$ regularization. On the other hand, if the number of noise predictors is negligible with respect to the number of classes, then the $L_1$ and $L_2$ methods are comparable.

The rest of the article is organized as follows. Section \ref{sec:review} provides some background information. The main methods are introduced in Section \ref{sec:method}. Section \ref{sec:implement} presents the algorithms and tuning parameter selection. A novel statistical learning theory is provided in Section \ref{sec:theory}. Section  \ref{sec:numerical} includes all the numerical studies. Some concluding remarks are given in Section \ref{sec:conclude}. Most technical proofs are collected in the Supplementary Materials.

\section{Background}\label{sec:review}
Let $P_j(\bx) = \pr(Y=j \mid \bX=\bx)$ be the class conditional  probability of observation $\bx$ for class $j$ ($j=\pm 1$ or $1,\dots,k$). In the binary case, it can be shown that the Bayes decision under the $0$-$d$-$1$ loss is, $\phi_{Bayes}(\bx)=+1$ if $P_{+1}(\bx) \ge 1-d$, $-1$ if $P_{-1}(\bx) \ge 1-d$, or $\rej$ otherwise \citep{herbei2006classification}.

Note that $\sum_{j}P_j(\bx)=1$ where $j=\pm 1$ for binary classification or $1,\dots,k$ for the multicategory case. Hence, for each $\bx$, $(P_j):=(P_j(\bx))_{j=\pm 1\textrm{ or }1,\dots,k}$ must fall on a simplex in $\R^k$. Throughout this article, we define the \textit{Bayes reject region} to be $R_{\textrm{Bayes}} := \{(P_j): \phi_{Bayes}(\bx) = \rej \}$, a region on this simplex. For example, in the binary case, we have $R_{\textrm{Bayes}} = \{(P_{+1},P_{-1}): d<P_{+1}(\bx)<1-d\}$.

While it is possible to achieve the reject option by first estimating the conditional probabilities $P_j(\bx)$ for each $\bx$ and then plugging the estimates in the Bayes rule (whose form in the multicategory case will be formally presented in Proposition \ref{prop1}), it is well known that probability estimation can be more difficult than mere label prediction \citep{WangShenLiu2008,Fuernkranz2010Preference,multiprob}, especially when the dimension $p$ is large \citep{refit}. Hence our goal here is to propose multicategory classifiers with a reject option that does not require explicit probability estimation.

We first briefly introduce the state-of-the-art for binary classification with a reject option. Section \ref{sec:reviewanglemc} reviews the angle-based multicategory classification methods.

\subsection{Binary Margin-based Classification with a Reject Option}\label{sec:reviewbinaryrej}
\begin{figure}[!htb]
\vspace{-2em}
\begin{center}
\includegraphics[angle=0,width=0.4\textwidth,totalheight=0.4\textwidth]{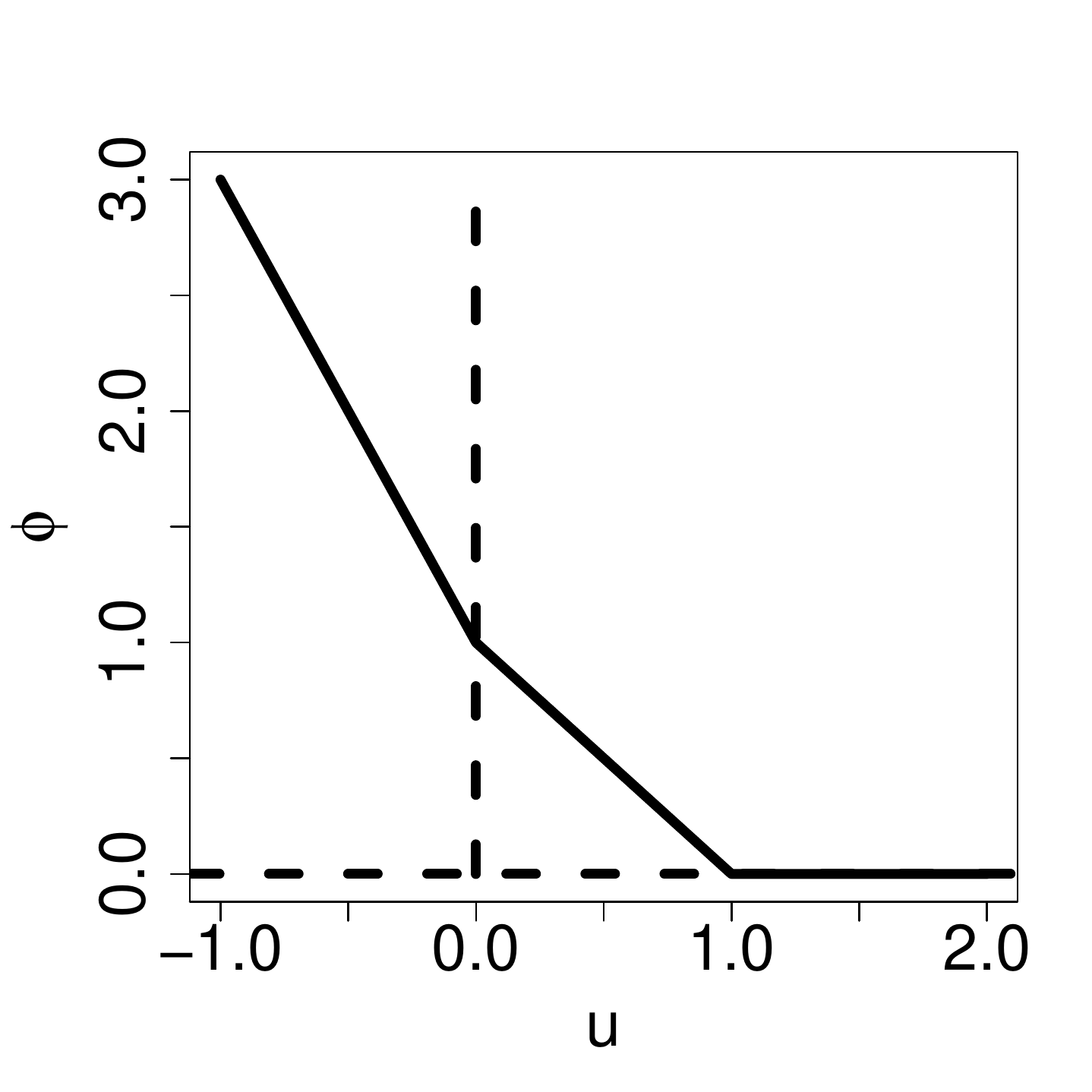}
\end{center}
\vspace{-2em}
\caption{{\small The modified hinge loss $\psi$ for binary problems \citep{bartlett2008classification}.
}}
\label{binarylossplot}
\end{figure}

The seminal paper of \citet{bartlett2008classification} proposed a novel method that employed a modified hinge loss $\psi$ for binary classification with a reject option. In particular, $\psi(u)=0$ if $u\ge 1$, $\psi(u)=1-u$ if $0\le u< 1$, and $\psi(u)=1-au$ otherwise, where $a = (1-d)/d>1$ (see Figure~\ref{binarylossplot}.) Define
$f^*(\bx) = \arginf_{f\in\mathcal{F}} \E [\psi\{Yf(\bX)\} \mid \bX=\bx]$ to be the minimizer of the conditional expected loss (for an appropriate space $\mathcal{F}$) and define the associated classifier to be $\phi_{f^*}(\bx)=\rej$ if $f^*(\bx)=0$, or $\textrm{sign}\{f^*(\bx)\}$ otherwise. Then the $f^*$-\textit{reject region} is defined as $R_{f^*} = \{(P_{+1},P_{-1}): f^*(\bx)=0 \}$. \citet{bartlett2008classification} showed that their $\phi_{f^*}$ coincided with the Bayes rule $\phi_{Bayes}$ and hence, $R_{\textrm{Bayes}} = R_{f^*}$.

\subsection{Angle-based Multicategory Classification}\label{sec:reviewanglemc}
\citet{MAC} showed that multicategory margin-based classification methods with $k$ classification functions under the sum-to-zero constraint can be inefficient, and proposed the angle-based classification framework. They showed that angle-based classifiers are competitive in terms of classification accuracy and computational speed, especially when $p$ is large. The idea of angle-based classifiers are briefly introduced here. For a problem with $k$ classes, consider a centered simplex in $\R^{k-1}$ with $k$ vertices, $\mY = \{\mY_1,\ldots,\mY_k\}$. Here
\begin{eqnarray*}
\mY_j=\left\{ \begin{array}{ll}
(k-1)^{-1/2}\bone_{k-1} & ~~~ j=1,\\
-(1+k^{1/2}) / \{(k-1)^{3/2}\} \bone_{k-1}+\{k/(k-1)\}^{1/2}\be_{j-1} & ~~~ 2 \le j \le k,
\end{array} \right.
\end{eqnarray*}
where $\bone_{k-1}\in\R^{k-1}$ is a vector of all $1$'s, and $\be_j\in\R^{k-1}$ has $1$ on its $j$th element and $0$ elsewhere. One can verify that $\mY_j$'s have unit norms, and the pairwise distances between $\mY_i$ and $\mY_j$ are the same for all $i \ne j \in \{1,\ldots,k\}$. Therefore, $\mY$ forms a simplex with $k$ vertices in $\mbbR^{k-1}$. We use $\mY_j$ as the surrogate coding vector for the class label `$j$'. In angle-based methods, a vector-valued classification function $\bdf$ maps $\bx$ to $\bdf(\bx)\in\R^{k-1}$. Each $\bdf(\bx)$ induces $k$ angles with $\mY_1,\ldots,\mY_k$, namely, $\angle(\mY_j, \bdf)$, $j=1,\ldots,k$. \citet{MAC} proposed to use the prediction rule $\phi(\bx) = \argmin_{j=1,\ldots,k} \angle(\mY_j, \bdf(\bx))= \argmax_{j=1,\ldots,k} \langle \mY_j,\bdf(\bx) \rangle$. Here, the inner product $\langle \mY_j,\bdf(\bx) \rangle$ can be viewed as an analog to the functional margin in a non-angle-based method, and hence is referred to as an angle margin hereafter. From this point of view, \citet{MAC} proposed to solve the following optimization problem to find $\bdf$ within some functional space $\mathcal{F}$,
\begin{align}
\min_{\bdf \in \mF}~~n^{-1} \sum_{i=1}^n \tau \{ \langle \mY_{y_i},\bdf(\bx_i) \rangle \}, ~\textrm{subject to } J(\bdf) \le s,
\label{dualloss}
\end{align}
where $\tau(\cdot)$ is a common binary margin-based surrogate loss function, $J(\bdf)$ is a penalty on $\bdf$ to prevent overfitting, and $s$ is a tuning parameter to balance the goodness of fit and the complexity of the model. The optimization (\ref{dualloss}) encourages a large value for $\langle \mY_{y_i},\bdf(\bx_i) \rangle$.

\section{Methodology}\label{sec:method}
In this section, we introduce our main methods, namely, a multicategory classifier with a reject option in Section \ref{sec:rejectoption}, and one with both reject and refine options in Section \ref{sec:refinementoption}.

\subsection{Multicategory Classification with a Reject Option}\label{sec:rejectoption}
Given an observation $\bx$, recall the definition of $P_j(\bx)$. Let $P_{(j)}(\bx)$ be the $j$th greatest value among $P_j(\bx)$'s, let $y_{(j)}$ be the class label corresponding to $P_{(j)}(\bx)$, and define $\mY_{(j)}$ to be the coding vector for $y_{(j)}$. Note that $y_{(j)}$ is not necessarily the true class label for $\bx$, but is its $j$th most plausible class.
Lastly, we define $Q_j=1-P_j$, and $Q_{(j)}=1-P_{(j)}$.

Our approach is inspired by the work of \citet{bartlett2008classification} for binary problems. In particular, their loss function was $\psi(u) = H(u) + (a-1) [-u]_+$, where $H(u):=[1-u]_+$ was the hinge loss function for SVM and $a-1= (1-d)/d-1>0$ was an additional slope added to the hinge loss for $u<0$. One can view $\psi$ as the hinge loss, \textit{bent} at $u=0$ so that the left derivative $-a$ is (negatively) larger than the right derivative $-1$. Denote the theoretical minimizer $f^*(\bx) = \arginf_{f\in\mathcal{F}} \E [\psi\{Yf(\bX)\} \mid \bX=\bx]$. The bent loss function can keep $f^*$ at $0$ if $P_j$ is not significantly different from $1-P_j$ ($j=+1,-1$), thus leading to a rejection in this case. In particular, $f^*$ is positive if $P_{+1}>1-d$, is negative if $P_{+1}<d$, and remains $0$ if $d < P_{+1} < 1-d $. Note that comparing $P_{+1}$ with $\{d,1-d\}$ is equivalent to comparing $Q_{+1}/Q_{-1}$ with $\{1/a,a\}$.

Inspired by these observations, to realize a reject option for multicategory classification, we employ a similar technique, namely, to use a \textit{bent} loss function that has different left and right derivatives at $0$. Specifically, we equip an angle-based multicategory classifier with a modified loss, with the aim to have the angle margin $\langle \mY_j,\bdf^*(\bx) \rangle = 0$ for all $j=1,\ldots,k$, where $\bdf^*(\bx)$ is the theoretical minimizer of the loss (to be defined more precisely later) if the class conditional probability $P_j(\bx)$'s are not significantly different from each other; note that this implies that $P_{(1)}$ is not large enough and that $Q_j$'s are similar as well. We will show in Proposition \ref{proposition2} that this is indeed the case.

\begin{figure}[!h]
\begin{center}
\subfigure[Loss function $\ell_{\textrm{SVM}}$ in (\ref{mainloss}) with $a = 2$.
]{\includegraphics[angle=0,width=0.4\textwidth,totalheight=0.4\textwidth]{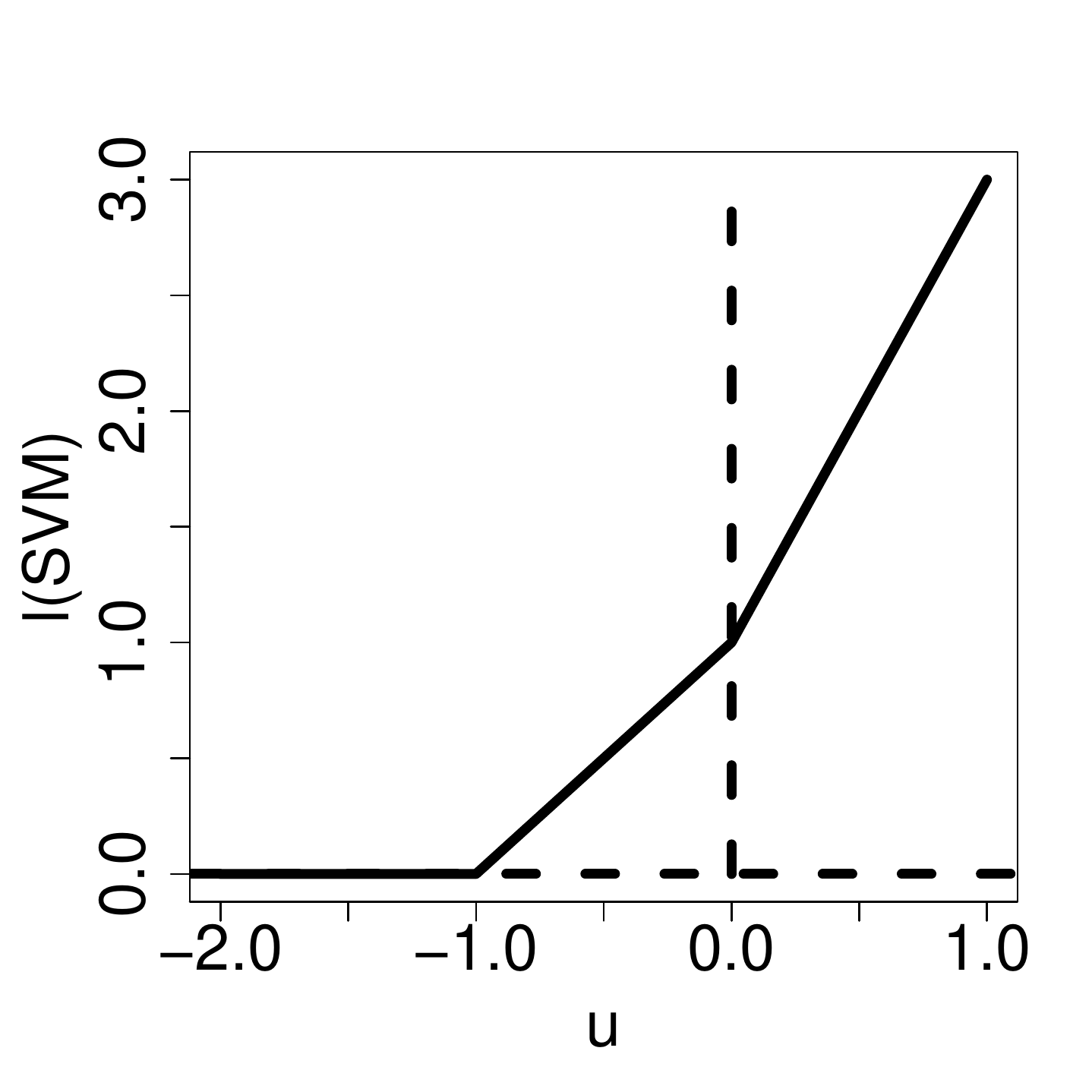}}
\subfigure[Loss function $\ell_{\textrm{DWD}}$ in (\ref{mainloss}) with $a = 2$.
]{\includegraphics[angle=0,width=0.4\textwidth,totalheight=0.4\textwidth]{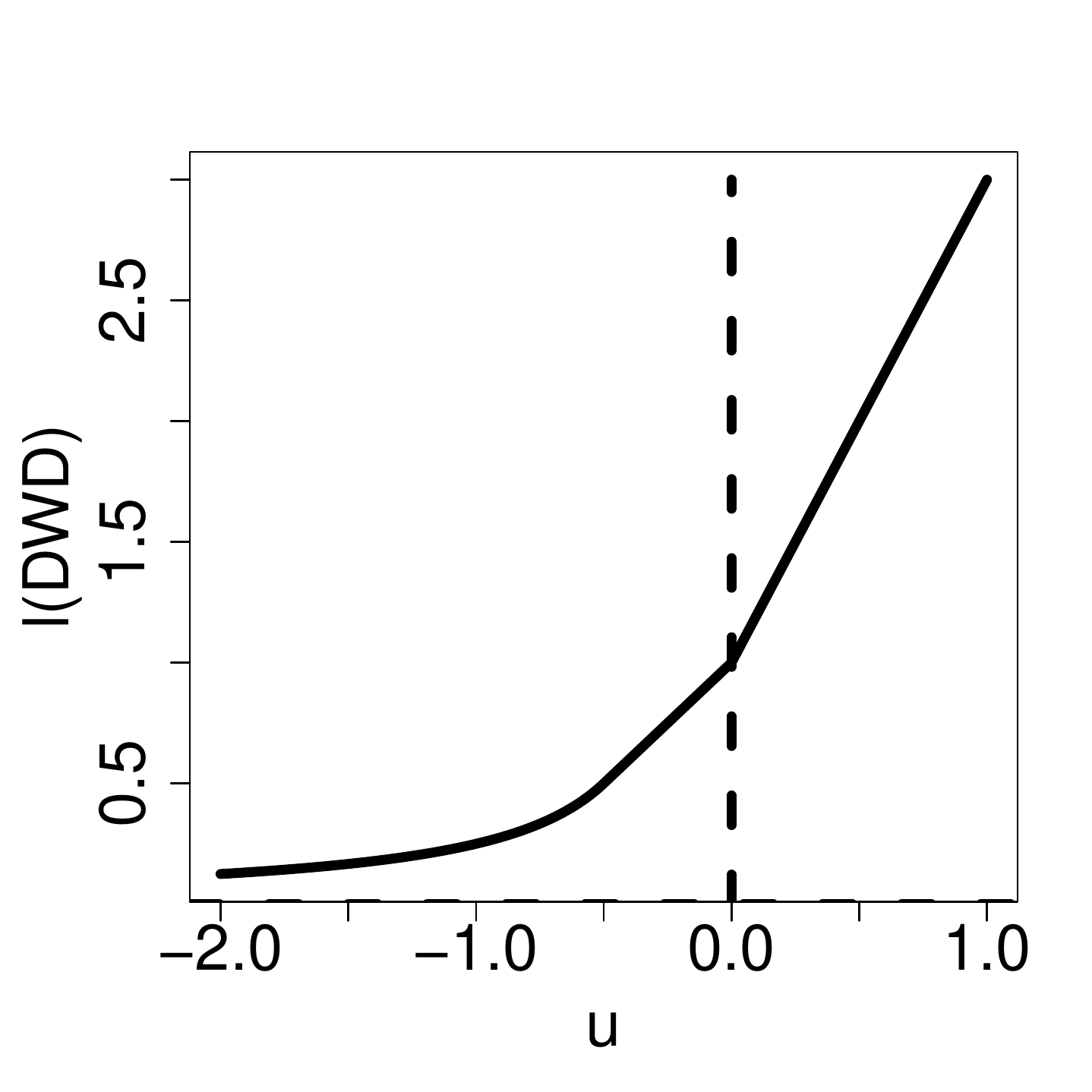}}
\end{center}
\caption{{\small
Plots of the bent loss functions for multicategory classification with a reject option.
}}
\label{multilossplot}
\end{figure}

For any observation $(\bx,y)$ and function $\bdf$, we propose a loss function defined as 
\begin{align}
\sum_{j \ne y} \ell\{ \langle \mY_j,\bdf(\bx) \rangle \}= \sum_{j \ne y} \Big[\ell_1\{ \langle \mY_j,\bdf(\bx) \rangle \}+\ell_2\{ \langle \mY_j,\bdf(\bx) \rangle \}\Big],
\label{mainloss}
\end{align}
where $\ell(u)=\ell_1(u)+\ell_2(u)$. Here $\ell_1(u) = \tau(-u)$ and $\tau$ is the loss function for any Fisher consistent binary margin-based classifier (such as the hinge loss, the DWD loss, the LUM loss and the FLAME loss.) Throughout this paper we assume $\ell_1'(0)=1$ for simplicity. Furthermore, $\ell_2(u)$ is defined so that $\ell'(u) \equiv a>1$ for $u>0$, and $\ell_2(u)=0$ for $u<0$. Hence $\ell$ is the result of bending $\ell_1$ using $\ell_2$. This will be illustrated in Figure \ref{multilossplot} using two typical loss functions. The loss function (\ref{mainloss}) is the sum of $\ell$ over all class $j$'s not equal to the true class $y$. With this loss function, our classification function  is obtained by,
\begin{align}
\hat\bdf=\argmin_{\bdf \in \mF}~n^{-1} \sum_{i=1}^n \sum_{j \ne y_i} \ell\{ \langle \mY_j,\bdf(\bx_i) \rangle \}, ~\textrm{subject to } J(\bdf) \le s.
\label{completeloss}
\end{align}
The monotonically increasing loss function $\ell$ encourages a small value of $\langle \mY_j,\bdf(\bx_i) \rangle$ for $j \ne y_i$ which indirectly maximizes $\langle \mY_{y_i},\bdf(\bx_i) \rangle$ since $\sum_{j=1}^k \mY_j= \boldsymbol{0}$.

With $\delta$ a small positive constant, define the soft thresholding operator \citep{Donoho1995De} as $S_{\delta}(c) = \textrm{sign}(c)\max(|c|-\delta,0)$. The induced classifier can be summarized as,
\begin{align}\label{predictionrule}
\phi_{\hat\bdf}(\bx)=\begin{cases}
\rej & \textrm{if }  S_{\delta}(\langle \mY_j, \hat\bdf \rangle)=0,~\forall j, \\
\argmax_{j=1,\ldots,k} \langle \mY_j, \hat\bdf(\bx) \rangle & \textrm{otherwise}.
\end{cases}
\end{align}
That is, we report a rejection when all $\langle \mY_j, \hat\bdf \rangle$'s are close to 0.

Our method is very general, as one can use any Fisher consistent binary margin-based loss and extend the binary classifier to the multicategory case, meanwhile allowing for a reject option. For the purpose of illustration, in this section we generalize two popular binary margin-based classifiers, SVM and DWD. The \textit{bent} SVM and DWD losses are,
$$\ell_{\textrm{SVM}}(u)=\left\{ \begin{array}{ll}
0 & \textrm{if } u < -1,\\
1+u & \textrm{if } -1 \le u < 0,\\
1+a u & \textrm{otherwise},
\end{array} \right.\makebox{and }
\ell_{\textrm{DWD}}(u)=\left\{ \begin{array}{ll}
-\frac{1}{4u} & \textrm{if } u < -0.5,\\
1+u & \textrm{if } -0.5 \le u < 0,\\
1+a u & \textrm{otherwise}.
\end{array} \right.
$$
We plot $\ell_{\textrm{SVM}}$ and $\ell_{\textrm{DWD}}$ in Figure~\ref{multilossplot}.

To provide more insights to the new classifier, we first study the population version of $\hat\bdf$, namely, the theoretical minimizer $\bdf^*$, and its associated reject region. We will compare the reject region of our method with the Bayes reject region under a generalized $0$-$d$-$1$ loss, and show that our methods mimic the latter, which helps to justify our approach from a theoretical view.

\begin{proposition}
Let $\ell$ be a bent loss function as defined in (\ref{mainloss}), with $\ell'(0-)=1$  and $\ell'(0+)=a>1$. For the sequence $Q_{(1)}\le Q_{(2)} \le \dots \le Q_{(k)}$, if there exists some $1\le s\le k-1$ such that $Q_{(s)}/Q_{(1)}< a$ and $Q_{(s+1)}/Q_{(1)}\ge a$, then the theoretical minimizer $\bdf^*$ of the conditional expected loss $\E \{ \sum_{j \ne Y} \ell\{ \langle \mY_j,\bdf(\bX) \rangle \} \mid \bX=\bx\}$ satisfies that $\langle \mY_{(1)},\bdf^*(\bx) \rangle >0$, $\langle \mY_{(2)},\bdf^*(\bx) \rangle = \dots = \langle \mY_{(s)},\bdf^*(\bx) \rangle = 0$, and $\langle \mY_{(t)} ,\bdf^*(\bx)\rangle <0$ for all $t\ge s+1$; otherwise, $\langle \mY_{(j)},\bdf^*(\bx) \rangle=0$ for all $j=1,\dots,k$.
\label{proposition2}
\end{proposition}

Proposition~\ref{proposition2} indicates that $\langle \mY_{(j)},\bdf^*(\bx) \rangle=0$ for all $j$ when $Q_{(k)}/Q_{(1)}<a$, that is, the class conditional probability of the most plausible class $y_{(1)}$ is not significantly different from that of the least plausible class $y_{(k)}$, by a ratio not exceeding $a>1$.

\begin{figure}[!t]
	\begin{center}
	\includegraphics[width=0.6\textwidth]{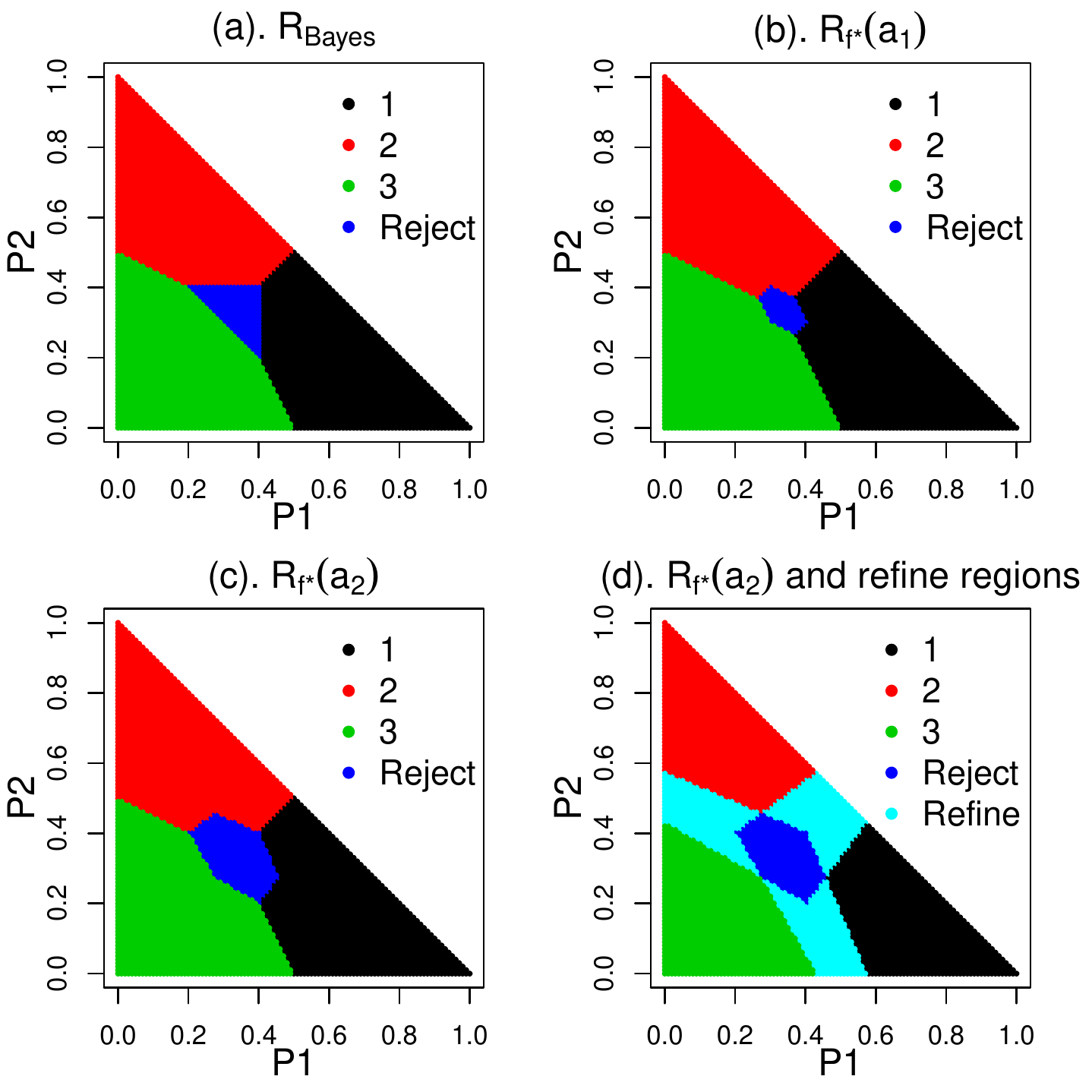}
	\end{center}
	\vspace{-1em}
	\caption{{\small
	(a) The Bayes reject region with the generalized $0$-$d$-$1$ loss ($k=3$ and $d=0.6$). (b)---(d) The $\bdf^*$-reject and refine regions with values of $a_1$ and $a_2$ defined in Proposition \ref{upperandlowerfora}.
	}}
	\label{rejectregion}
\end{figure}

Hence, the corresponding $\bdf^*$-reject region is $R_{\bdf^*}(a)= \{(P_1,\ldots,P_k): \langle \mY_j, \bdf^* \rangle=0,~\forall j\}= \{(P_1,\ldots,P_k): Q_{(k)} < aQ_{(1)}\}$ which depends on the parameter $a$. When the context is clear, we may use notation $R_{\bdf^*}$ without explicitly declaring its dependence on $a$. On Panels (b) and (c) of Figure~\ref{rejectregion}, we plot $R_{\bdf^*}(a)$ for a three-class example with two values of $a$, $a_1$ and $a_2$, defined in Proposition \ref{upperandlowerfora}.

For each $a$, the $\bdf^*$-reject region is near the center of the simplex, which is where the class conditional probability $P_j$'s are close to each other. Intuitively, that is a difficult observation to classify. Next, consider a natural generalization of the (binary) $0$-$d$-$1$ loss in \citet{herbei2006classification} to the multicategory case, which assigns $0$ for correct decisions, $1$ for mistakes, and $d$ for $\rej$. In a $k$-class problem, we must have $0< d \le (k-1)/k$ to prevent the reject option from being inadmissible. The next proposition gives the Bayes classifier under the generalized $0$-$d$-$1$ loss for multicategory classification, which depends on $P_{(1)}$ only. The Bayes reject region is $R_{\textrm{Bayes}} = \{(P_1,\ldots,P_k): P_{(1)}\le1-d \}$ (see Panel (a) of Figure~\ref{rejectregion}.)
\begin{proposition}\label{prop1}
\citep{chow1970optimum} For the $0$-$d$-$1$ loss in multicategory classification, the Bayes classifier is $\phi_{\textrm{Bayes}}(\bx)=y_{(1)}$ if $P_{(1)}(\bx) > 1-d$, and $\rej$ otherwise.
\end{proposition}

One would expect a good classifier with a reject option to have a reject region that resembles (or even coincides with) that of the Bayes rule (under an appropriate loss function). Indeed, one can deduct from Proposition~\ref{proposition2} that for any Fisher consistent binary loss function with $k=2$ and $a=(1-d)/d$, our $R_{\bdf^*}$ coincides with the Bayes reject region $R_{\textrm{Bayes}}$  under the $0$-$d$-$1$ loss. However, in the multicategory case, this property generally does not hold. The next proposition gives the greatest $a_1$ and smallest $a_2$ such that $R_{\bdf^*}(a_1)$ and $R_{\bdf^*}(a_2)$ bound $R_{\textrm{Bayes}}$ from two sides.

\begin{proposition}
For a $k$-class problem with the cost for rejection $d$, define $a_1 = (k-1-d)/(kd-d)$ and $a_2 = (k-1)(1-d)/d$. Then we have $R_{\bdf^*}(a_1) \subset R_{\textrm{Bayes}} \subset R_{\bdf^*}(a_2)$. The bounds are tight in the sense that for any $a$ such that $a_1<a<a_2$, $R_{\bdf^*}(a) \not\subset R_{\textrm{Bayes}}$ and $R_{\textrm{Bayes}} \not\subset R_{\bdf^*}(a) $.
\label{upperandlowerfora}
\end{proposition}

Panels (b) and (c) in Figure \ref{rejectregion} show the $\bdf^*$-reject regions for $a_1$ and $a_2$. From the comparison between these two reject regions and the Bayes reject region shown in Panel (a), one can see that our method induces a reject region that closely approximates the Bayes reject region. In practice, one can choose $a$ from $[a_1,a_2]$ for such an approximation. The issue of tuning the parameter $a$ is deferred to Section \ref{sec:tuning}.

In each panel among (a), (b) and (c), the reject region occupies the center of the simplex where all $P_j$ are close to each other (\textit{i.e.} $P_{(1)}$ is not large enough). Out of that area, some or all the classes other than the dominating class $y_{(1)}$ would appear to be \textit{unlikely} and hence are ruled out. In this case, a rejection is \textit{not} yielded by (\ref{predictionrule}).

\subsection{Classification with a Refinement Option}\label{sec:refinementoption}

The previous subsection is built on the assumption that a reject option is necessary when an observation falls into the reject region, depicted in Figure \ref{rejectregion}, where all classes seem to be equally likely and it is difficult to distinguish one class from another. On the other hand, even if an observation is \textit{not} in the reject region, it is not necessarily the case that a definite classification is desirable. This is the main point of the current subsection. In each of Panels (a)-(c) of Figure \ref{rejectregion}, out of the blue reject region, there are still areas where some confusion may occur between two classes. For example, many observations near the boundary between the black (class 1) and the red (class 2) regions are not likely to be from class 3, but we still have difficulty determining between class 1 and class 2. A method which is only capable of yielding rejections is still not able to effectively avoid an expensive misclassification which is very likely to happen in this situation. This naturally motivates a new \textit{refine} option for multicategory classification, in which, we may rule out class 3 and predict the observation to be from either class 1 or class 2. On one hand, we can avoid a potential misclassification by using a set of classes as the prediction; on the other hand, the set prediction provides additional information compared to what a rejection would do (which is almost null.)

The discussion above suggests that the complement of the reject region (the previous \textit{definite} regions) be further partitioned to some \textit{definite} regions and \textit{refine} regions. In Figure \ref{rejectregion}, for example, in addition to rejections, we should have (a) three definite regions where the prediction is a single class label, 1, 2 or 3, and (b) three refine regions where the prediction is a set of two classes, namely, $\{1,2\}$, $\{2,3\}$ or $\{1,3\}$.

To this end, we review the results of Proposition \ref{proposition2}: a rejection occurs (all the angle margins $\langle \mathcal{Y}_{(s)},\bdf^*(\bx)\rangle=0$) when the most plausible class $y_{(1)}$ cannot be distinguished from the least plausible one $y_{(k)}$ (since $Q_{(k)}<aQ_{(1)}$); otherwise, the angle margin $\langle \mathcal{Y}_{(1)},\bdf^*(\bx)\rangle$ for the most plausible class $y_{(1)}$ is positive, the angle margins for some less plausible classes $s$ are zero, although the conditional probabilities of these classes are still close to that of $y_{(1)}$ (since $Q_{s}<aQ_{(1)}$), and the angle margins for the implausible classes are all negative. Hence we may use the angle margins to define predictions, since they reflect the plausibility of a class label for an observation. A general guideline is that a positively large angle margin suggests a label prediction, the presence of some angle margins close to 0 and some angle margins negatively large suggests refinement (and ruling out those implausible), and the case of all angle margin close to 0 indicates rejection.

In reality, since the empirical angle margin $\langle \mathcal{Y}_{j},\hat\bdf(\bx)\rangle$ may be deviated from the theoretical counterpart $\langle \mathcal{Y}_{j},\bdf^*(\bx)\rangle$ for a finite sample problem, the gap between angle margins may not appear obvious. In this case, we employ a soft-thresholding technique to distinguish significantly large and small angle margins. In particular, with the thresholded angle margins, our new classifier with both reject and refine options is defined as,
\begin{align}\allowdisplaybreaks\label{refinementoptionrule}
\phi_{\hat{\bdf}}^{\textrm{set}}=\begin{cases}
\rej & \textrm{if }  S_{\delta}(\langle \mY_j, \hat\bdf \rangle)=0,~\forall j, \\
\{j: S_{\delta}(\langle \mY_j, \hat\bdf \rangle)>0\} & \textrm{if } S_{\delta}(\langle \mY_j, \hat\bdf \rangle)>0,~\textrm{for some }j, \\
\{j:S_{\delta}(\langle \mY_j, \hat\bdf \rangle)=0\} & \textrm{otherwise}.
\end{cases}
\end{align}
Note that the reject rule, the first line in (\ref{refinementoptionrule}), is identical to that in (\ref{predictionrule}). This corresponds to the case that all angle margins are close to 0, implying that all the class conditional probabilities are close to each other. The second line attempts to find the most significantly large margin, and hence the most plausible class. In our numerical experience, we occasionally observe cases with multiple significantly large margins which are close to each other. In this case, we have chosen to include all those plausible classes (if any) as a set prediction. The third line corresponds to the case where the most plausible class is \textit{not} significantly different from some other classes and we resort to ruling out those implausible classes (those with significantly negatively large margins) instead.

It can be seen that the union of the second and third cases in (\ref{refinementoptionrule}) is identical to the definite label prediction region in (\ref{predictionrule}) (the second line therein). However, when an observation belongs to the third case in (\ref{refinementoptionrule}), the rule in (\ref{predictionrule}) recklessly reports a single label as the prediction, while the novel refinement rule (\ref{refinementoptionrule}) here uses a set prediction. This is the main difference between the classifiers in (\ref{predictionrule}) and (\ref{refinementoptionrule}).

For illustration, we plot the reject, refine and definite regions for a three-class problem on Panel (d) of Figure~\ref{rejectregion} for $a_2$. It can be seen that the three \textit{refine} (cyan) regions are cut from the previous \textit{definite} regions in Panel (c) and hence the current \textit{definite} regions are smaller than in (c) as well. More importantly, one may hold more confidence for a label prediction made by the new classifier (\ref{refinementoptionrule}). In Section~\ref{sec:numerical}, we demonstrate through numerical examples that the classification accuracy on the refine region in (\ref{refinementoptionrule}) can be significantly improved, compared to the classification accuracy on the counterpart of (\ref{predictionrule}).

In both (\ref{predictionrule}) and (\ref{refinementoptionrule}), the choice of $\delta$ is a matter of tuning parameter. The details of tuning $\delta$ are given in the next section.

\section{Optimization and Tuning Parameter Selection}\label{sec:implement}

In this section, we discuss how to solve the optimization problem (\ref{completeloss}), from which both our methods (\ref{predictionrule}) and (\ref{refinementoptionrule}) are derived. Various approaches are possible, depending on the choice of $\ell$, $\mF$ and $J(\bdf)$. For demonstration purpose, in this section we use the reversed hinge loss for $\ell_1$. We let $J(\bdf)$ be the  $L_2$ norm penalty in linear learning, and the squared norm penalty in kernel learning \citep{Schlkopf2002,kernelbook}. For other cases with some general properties, such as one with a differentiable $\ell_1$ and a separable penalty function, one can solve (\ref{completeloss}) by the alternating direction method of multipliers \citep{boyd2011distributed}. We have developed fast implementations for our methods based on the hinge loss, the DWD loss and the Soft classifier loss \citep{LUM}. These algorithms will be publicly available in R.

\subsection{Optimization}\label{optimization}
We start our discussion from linear learning. Suppose $f_q(\bx) = \bx^T \bbeta_q$ for $q=1,\ldots,k-1$. Notice that we include the intercept terms in the $\bbeta_q$'s by catenating $1$ to $\bx$. The $L_2$ penalty $J(\bdf)$ can be written as $J(\bdf) = \sum_{q=1}^{k-1} \bbeta_q^T \bbeta_q$. The bent hinge loss $\ell$ can be decomposed as $\ell(u) = [1+u]_+ + (a-1)[u]_+$. After a series of introduction of Lagrangian multiplier and slack variables, and manipulations due to the KKT conditions (detailed derivations of the algorithms can be found in the Supplementary Materials), we can show that the optimization problem (\ref{completeloss}) is equivalent to
\begin{align}\label{linearoptimization}
\min_{\alpha_{ij}, \gamma_{ij}} & ~~ \frac{n\lambda}{2} \sum_{q=1}^{k-1} \bbeta_q^T \bbeta_q - \sum_{i=1}^n \sum_{j \ne y_i}  \alpha_{ij}, \nonumber \\
\textrm{subject to}& ~~ 0 \le \alpha_{ij} \le A_{ij}, ~ 0 \le \gamma_{ij} \le A_{ij},~ i=1,\ldots,n,~j = 1,\ldots, k,
\end{align}
where $\bbeta_q = -\frac{1}{n \lambda} \sum_{i=1}^n \sum_{j \ne y_i} \{ \alpha_{ij} + (a-1) \gamma_{ij} \} \mY_{j,q} \bx_i.$ Observe that the objective function is quadratic in terms of $\alpha_{ij}$'s and $\gamma_{ij}$'s, and the constraints are box constraints. Therefore, one can solve (\ref{linearoptimization}) via the very fast coordinate descent algorithm \citep{Hastie10}. Moreover, as the objective function is quadratic, for each coordinate-wise update, the solution can be explicitly calculated. This greatly boosts the computational speed.

Similarly, for kernel learning, we can use $f_q(\bx) = \sum_{i=1}^n K(\bx_i,\bx) \theta_{q,i} + \theta_{q,0}$, $q=1,\ldots,k-1$, for kernel function $K(\cdot,\cdot)$, where the square norm penalty is $\sum_{q=1}^{k-1} \btheta_q^T \bK \btheta_q$, and $\theta_{q,i}$ is the $i$th element of $\btheta_q$. In the same manner as above, one can derive a fast solution to this problem.

\subsection{Tuning Parameter Selection}\label{sec:tuning}
There are three tuning parameters in our methods, namely $a$, $s$ and $\delta$. Here $a$ is associated with the cost of rejection $d$, where the latter should be fixed \textit{a priori}. In the numerical study, we find that the choice of $a$ does not affect the result much, as long as $a_1<a<a_2$. We recommend to try both $a_1$ and $a_2$ and use the one with a better result.

Parameter $s$ restricts the model space that the classifier is searched from. Typically $s$ is tuned from a grid of many candidate values. The optimal $s$ is chosen for one that minimizes the $0$-$d$-$1$ loss for a separate tuning data set or via cross-validation.

Lastly, $\delta>0$ is a small positive constant used to distinguish significantly large and small angle margins. Similar to $s$, we tune $\delta$ by choosing the one that leads to the smallest $0$-$d$-$1$ loss for a separate tuning data set or via cross-validation. However, note that solving the optimization problem (\ref{completeloss}) to obtain $\hat\bdf$ does not involve $\delta$; only the conversion from $\hat\bdf$ to the classifier $\phi_{\hat\bdf}$ or $\phi_{\hat\bdf}^{\textrm{set}}$ does. Hence tuning $\delta$ hardly adds to the computational cost.

\section{Statistical Learning Theory}\label{sec:theory}
In this section, we first study the convergence rate of the excess $\ell$-risk under various settings. In particular, we study the cases of linear learning with $L_1$ and $L_2$ penalties, and kernel learning with the squared norm penalty. Then, we improve our results with an additional low noise assumption, analogous to Tsybakov's margin condition \citep{tsybakov2004optimal}.

\subsection{General Convergence Rate of the Excess $\ell$-Risk}\label{generalrates}
In the literature, the excess $\ell$-risk for a learning procedure has been studied by many authors in different settings. See \citet{zhangtong04} and \citet{Bartlett06} for standard binary classification, \citet{multipsi06}, \citet{Wang2007}, and \citet{MAC} for multicategory classification, and \citet{herbei2006classification}, \citet{wegkamp2007lasso}, and \citet{wegkamp2011support} for binary classification with a reject option. We focus on the excess $\ell$-risk for the multicategory classification with a reject option.

We first consider linear learning with a diverging number of predictors $p$ and a diverging number of classes $k$. In the statistical learning literature, it is becoming increasingly popular to consider large $p$ as $n \rightarrow \infty$ \citep[for example,][among others.]{fan2008sure,mai2012kolmogorov,cai2012estimating} On the other hand, for classification problems, not much attention has been paid to the large $k$ situation. Recently, \citet{gupta2014training} studied classification problems with tens of thousands of classes. However, the theoretical property of classifiers with diverging $k$ remains largely unknown.

First, we assume that each predictor is bounded within $[0,1]$, though our theory can be generalized to cases where it is uniformly bounded. As the number of predictors $p$ and the number of classes $k$ diverge, we let the underlying distribution $\mbbP(\bx,y)$ be defined on $\big([0,1]^{\infty} \times \{1,\ldots,k,\ldots,\}, \sigma^{\infty} ([0,1]^{\infty}) \times 2^{ \{1,\ldots,k,\ldots,\} } \big)$, where $\sigma^{\infty}([0,1]^{\infty})$ is the $\sigma$-field generated by open balls with the topology under the uniform metric $d(\bx,\bx') = \sup_{l=1,\dots,k,\dots} | x_l -  x'_l|$, and $2^{ \{1,\ldots,k,\ldots,\}}$ is the power set of $\{1,\ldots,k,\ldots,\}$ and hence a $\sigma$-field.

For linear learning, we have $\bdf = (f_1,\ldots,f_{k-1})^T$ with $f_q(\bx) = \bbeta_q^T \bx$, $q=1,\ldots,k-1$. We define $\mF(p,k,s) = \{ \bdf = (f_1,\ldots,f_{k-1})^T: f_q(\bx) = \bbeta_q^T \bx,~q=1,\ldots,k-1,~ J(\bdf) \le s\}$. For the $L_1$ penalty, $J(\bdf)= \sum_{q=1}^{k-1} \|\bbeta_q\|_1$, and for the $L_2$ penalty, $J(\bdf)= \sum_{q=1}^{k-1} \|\bbeta_q\|_2^2$. Let $\mF(p,k) = \bigcup_{0 \le s < \infty} \mF(p,k,s)$ be the full $p$-dimensional model with $k$ classes. Recall that $\hat{\bdf} = \argmin_{\bdf \in \mF(p,k,s)} \frac{1}{n} \sum_{i=1}^n \sum_{j \ne y_i} \ell\{\langle \bdf(\bx_i), \mY_j \rangle\}.$ Let the best classification function be denoted by $\bdf^{(p,k)} = \arginf_{\bdf \in \mF(p,k)} \E[\sum_{j \ne y} \ell \{ \langle \bdf(\bx), \mY_j \rangle \}]$.

For any classification function $\bdf$, the excess $\ell$-risk $e(\bdf, \bdf^{(p,k)})$ is defined as $$e(\bdf, \bdf^{(p,k)})=\E[\sum_{j \ne Y} \ell \{ \langle \bdf(\bX), \mY_j \rangle \}] - \E[\sum_{j \ne Y} \ell \{ \langle \bdf^{(p,k)}(\bX), \mY_j \rangle \}].$$ We denote $d_{n,p,k} = \inf_{\bdf \in \mF(p,k,s)} e_{\ell} (\bdf, \bdf^{(p,k)})$ as the approximation error between $\mF(p,k,s)$ and $\mF(p,k)$. Theorem \ref{excessellriskthm} establishes the convergence rate of $e(\hat{\bdf}, \bdf^{(p,k)})$ as $n,p,k \rightarrow \infty$.

\begin{theorem}
Assume $r=\{ \log(p k) / n \}^{1/2} \rightarrow 0$ as $n,p,k \rightarrow \infty$. For linear learning with the $L_1$ penalty, $e_{\ell} (\hat{\bdf}, \bdf^{(p,k)}) = O[\max \{ s k r \log(r^{-1}), d_{n,p,k}\}]$, almost surely under $\mbbP$. For the $L_2$ penalty, $e_{\ell} (\hat{\bdf}, \bdf^{(p,k)}) = O[\max\{ (p s)^{1/2} k r \log(r^{-1}), d_{n,p,k} \}]$, almost surely under $\mbbP$.
\label{excessellriskthm}
\end{theorem}

In Theorem~\ref{excessellriskthm}, $s$ controls the balance between the estimation error, that is $s k r \log(r^{-1})$ or $(p s)^{1/2} k r \log(r^{-1})$, and the approximation error $d_{n,p,k}$. As $s$ increases, $d_{n,p,k}$ decreases. The best trade off is one such that $s k r \log(r^{-1}) \sim d_{n,p,k}$ for the $L_1$ penalty, and $(p s)^{1/2} k r \log(r^{-1}) \sim d_{n,p,k}$ for the $L_2$ penalty. The convergence of the excess $\ell$-risk requires that $k =o(n^{1/2})$ and $\log(p) = o(n)$ for the $L_1$ penalized method, and $k =o(n^{1/2})$ and $p = o(n)$ for the $L_2$ method.

Theorem \ref{excessellriskthm} suggests that classification with a large number of classes can be very difficult. This helps to shed some light on the usefulness of our refine option. In particular, if a set of class labels frequently appears in set predictions (for instance, see Examples 2 and 3 in Section \ref{sec:numerical}), one can consider a refined classification problem (with labels restricted in the prediction set) and use a richer functional space $\mF$ if desired. Theorem \ref{excessellriskthm} suggests that the new classifier can have better performance since the number of classes is smaller.

When $k$ is bounded, and the classification signal is sparse, Theorem~\ref{excessellriskthm} demonstrates the effectiveness of the $L_1$ method: it can be verified that if the true classification signal is sparse, then one can choose a large enough but fixed $s$, such that the approximation error is $0$. In other words, $\bdf^{(p,k)} \in \mF(p,k,s)$ for some $s < \infty$. In this case, Theorem~\ref{excessellriskthm} can be greatly simplified.

\begin{corollary}
Assume that $k$ is bounded, and the true classification signal depends on finitely many predictors. Assume $r' = \{ \log(p) / n \}^{1/2}\rightarrow 0$ as $n,p \rightarrow \infty$. We can choose $s=s^*$ for all large $n$, such that $d_{n,p,k} = 0$. Consequently, for the $L_1$ penalty, $e_{\ell} (\hat{\bdf}, \bdf^{(p,k)}) = O \{ r' \log(r'^{-1})\}$, almost surely under $\mbbP$, and for the $L_2$ penalty, $e_{\ell} (\hat{\bdf}, \bdf^{(p,k)}) = O \{ p^{1/2} r' \log(r'^{-1}) \}$, almost surely under $\mbbP$.
\label{sparsecorollary}
\end{corollary}

On the other hand, for $k \rightarrow \infty$ as $n \rightarrow \infty$, we cannot have a fixed $s$ such that $d_{n,p,k} = 0$, even if the dimensionality $p$ is bounded. The next corollary considers a special situation where the number of true signal grows linearly with the number of classes. In this case, we can let $s=O(k)$, such that the approximation error is zero.

\begin{corollary}
Consider any classification sub-problem where the label $y$ is restricted in $\{1,\ldots,k_0\}$, for any $1<k_0<k$. Suppose that the classification signal for the restricted sub-problem depends on at most $c k_0$ predictors, where $c$ is a fixed positive integer that is universal for all $k_0$. Then for the complete problem with $k \rightarrow \infty$ classes,  one can choose $s = C k$ with a fixed constant $C>0$, such that the approximation error $d_{n,p,k} = 0$. Consequently, $e_{\ell} (\hat{\bdf}, \bdf^{(p,k)}) = O \{ k^2 r \log(r^{-1})\}$ for the $L_1$ penalty, and $O \{ p^{1/2} k^{3/2} r \log(r^{-1})\}$ for the $L_2$ penalty, almost surely under $\mbbP$.
\label{kcorollary}
\end{corollary}

A common scenario in which the assumptions of Corollary~\ref{kcorollary} hold is when each class has its own identifying attributes, and the number of signature attributes for each class is uniformly bounded by $c$. For instance, in cancer research, one may identify each cancer subtype with mutations on a small and non-overlapping group of feature genes. In this case, we can choose $s$ as a linear function of $k$ such that the approximation error is $0$. Another insight of Corollary~\ref{kcorollary} is that when there is no noise variable, that is, when $p=O(k)$, we have that the performance of the $L_2$ and $L_1$ regularization methods is comparable since the corresponding estimation errors have the same convergence rate.

Next, we study the convergence rate of the excess $\ell$-risk for kernel learning. To this end, we impose an assumption that the kernel is separable, and its corresponding kernel function is uniformly upper bounded. In other words, $K(\cdot,\cdot) < \infty$. \citet{Ingo07} and \citet{Blanchard2008}, among others, used a similar assumption.

For kernel learning with the squared norm penalty, recall from Section~\ref{optimization} that the estimated classification functions are of the form $f_q(\bx) = \sum_{i=1}^n \theta_{q, i} K(\bx_i,\bx) + \theta_{q, 0} $ with $q=1,\ldots, k-1$. We define $\mF(p,k,s) = \{ \bdf = (f_1,\ldots,f_{k-1})^T: f_q = \sum_{i=1}^n \theta_{q, i} K(\bx_i,\bx) + \theta_{q, 0}, \ J(\bdf) \le s\}$, where $J(\bdf) = \sum_{q=1}^{k-1} \btheta_q^T \bK \btheta_q + \sum_{q=1}^{k-1} \theta_{q,0}^2$. Note that the intercepts are included in the penalty. In the RKHS learning literature, many theoretical results are derived without the intercept term \citep{bousquet2002stability,chen2004support,steinwart2008support}. Our theory can incorporate regularized intercepts in the classification functions, hence is more general. Let $\mF(p,k)$, $\bdf^{(p,k)}$ and $e(\bdf, \bdf^{(p,k)})$ be defined analogously as in the linear learning case. The next theorem gives the convergence rate of the excess $\ell$-risk for kernel learning.

\begin{theorem}\label{excessellriskthm_kernel}
Assume $r=\{ \log(k) / n \}^{1/2} \rightarrow 0$ as $n,k \rightarrow \infty$. For RKHS learning, assume that the kernel is separable, and the corresponding kernel function is uniformly upper bounded. We then have $e_{\ell} (\hat{\bdf}, \bdf^{(p,k)}) = O[\max \{ s k r \log(r^{-1}), d_{n,p,k}\}]$, almost surely under $\mbbP$.
\end{theorem}

In Theorem~\ref{excessellriskthm_kernel}, the dimension of the predictors $p$ does not directly affect the estimation error $s k r \log(r^{-1})$. Instead, it is implicitly involved in the approximation error $d_{n,p,k}$. This is because the proof of Theorem~\ref{excessellriskthm_kernel} relies on the complexity of the function space $\mF(p,k,s)$, in terms of its covering number \citep{EPbook}. Note that Theorem~\ref{excessellriskthm_kernel} requires only that the kernel is separable and the kernel function is upper bounded, hence can be very general. On the other hand, if we restrict our consideration on a specific kernel, then more refined results can be obtained. For instance, many theoretical properties of the well known Gaussian kernel have been established. In \citet{zhou2002covering} and \citet{Ingo07}, the relation between the covering number of the corresponding function space and $p$ has been obtained. Therefore, one can modify the proof of Theorem~\ref{excessellriskthm_kernel} and explore the explicit effect of $p$ on the estimation error accordingly.

So far, we have obtained the convergence rate of the estimation error for our classifiers. For linear learning and kernel learning, the rate can be close to the \textit{parametric rate} $O(n^{-1/2})$, if $p$ and $k$ are negligible as $n \rightarrow \infty$. In the next section, we consider stronger assumptions, including a low noise assumption for multicategory classification problems. We show that faster rates are possible under these additional conditions.

\subsection{Fast Rate under Low Noise Assumption}\label{fast_rate_section}

In the literature, many theoretical results have been established for binary SVMs with assumptions similar to Tsybakov's margin condition \citep[see][and the references therein.]{Ingo07,bartlett2008classification,wegkamp2011support,zhao2012estimating} In this paper, we consider the margin condition in multicategory problems with a reject option for a general loss function $\ell_1$ in (\ref{mainloss}). We show that when the classification function is in certain RKHSs, for example the Gaussian kernel space, a faster rate of convergence of the excess $\ell$-risk can be obtained.

\begin{assumption}\label{lownoiseassumption}
(Low noise assumption) For $k$-class classification problems, we say that the distribution $\mbbP(\bx,y)$ satisfies the margin condition at threshold level $a$ with exponent $\alpha>0$, if there exists a constant $c \ge 1$ such that for all $t>0$,\begin{align}\label{lownoise}
\textrm{pr} \{|a(1-P_{(1)}) - (1-P_{(k)})|< t \} \le ct^{\alpha}.
\end{align}
\end{assumption}
Intuitively, under Assumption 1 with large $\alpha$, little probability mass is put around the boundary between the reject region and its complement. Thus, the classification signal is strong, and we expect that the estimation error can have a faster convergence rate. For binary SVM with a reject option, (\ref{lownoise}) reduces to the low noise assumption introduced in \citet{bartlett2008classification} with $a=(1-d)/d$.

Because we intend to consider a general loss function, we impose some minor restrictions on the loss and some assumptions on the distribution. The next assumption is needed to prevent $|\langle\mY_j, \bdf^*\rangle|$ from being too large, which yields a lower bound for the second order derivative of $\ell_1$ at the theoretical minimizer $\bdf^*$.

\begin{assumption}
The loss function $\ell_1(u)$ in (\ref{mainloss}) is twice differentiable for $u<0$. Furthermore, for any $\bx\in[0,1]^\infty$, the class conditional probability for any class $j$ is bounded away from $0$. In other words, $P_j(\bx) \ge \eta_0$ for a small and positive $\eta_0$.
\label{assumption2}
\end{assumption}

Theorem \ref{fast_rate_thm} improves the convergence rate under the new assumptions.

\begin{theorem}\label{fast_rate_thm}
Suppose Assumptions~\ref{lownoiseassumption} and~\ref{assumption2} hold with exponent $\alpha$. Then for our proposed method (\ref{completeloss}) with the Gaussian kernel, we have $e_{\ell} (\hat{\bdf}, \bdf^*) = O[\max \{ s k n^{-(1+\alpha)/(2+\alpha)}, d_{n,p,k}\}]$.
\end{theorem}
Hence, the estimation error can converge at a rate faster than $n^{-1/2}$. In particular, for a problem with fixed $k$ and $\bdf^{(p,k)} \in \mF(p,k,s)$ for a non-diverging $s$, if $\alpha \rightarrow \infty$, then the rate can become arbitrarily close to $n^{-1}$.

We remark that for a differentiable loss function $\ell_1$ whose derivative is strictly positive for small $u$, one may have $\langle\bdf^*, \mY_{(1)}\rangle \rightarrow \infty$ if some $P_j(\bx)$ goes to zero. In this perspective, Assumption~\ref{assumption2} helps to bound $\langle\bdf^*, \mY_{(1)}\rangle$. However, Assumption~\ref{assumption2} may not be needed for some special loss functions. For example, if $\ell_1$ is the reversed hinge loss, or the reversed FLAME loss proposed by \citet{qiao2015flexible}, we can drop Assumption~\ref{assumption2} while the result in Theorem~\ref{fast_rate_thm} remains valid. In general, if the loss function $\ell_1$ is flat for small enough $u$, we can remove Assumption~\ref{assumption2} from Theorem~\ref{fast_rate_thm}. See the proof and discussion of Theorem~\ref{fast_rate_thm} in the Supplementary Materials for more discussions.

\section{Numerical Studies}\label{sec:numerical}

In this section, we study the numerical performance of our proposed classifiers (one with a reject option only, and one with both reject and refinement options.) For classification problems with weak signals, we show that the empirical $0$-$d$-$1$ loss for classifiers with a reject option can be smaller than that for regular classifiers. Furthermore, we show that the refine option can often provide refined set prediction with very high accuracy. Due to its reliable performance, in practical problems, the refinement option can be used to identify classes that are highly confusable with each other, so that future tests can be dedicated to these classes for potential improvement in classification accuracy.

\subsection{Method of Comparisons}
\begin{figure}[htb]
\vspace{-2em}
\begin{center}
\includegraphics[angle=0,width=0.4\textwidth,totalheight=0.4\textwidth]{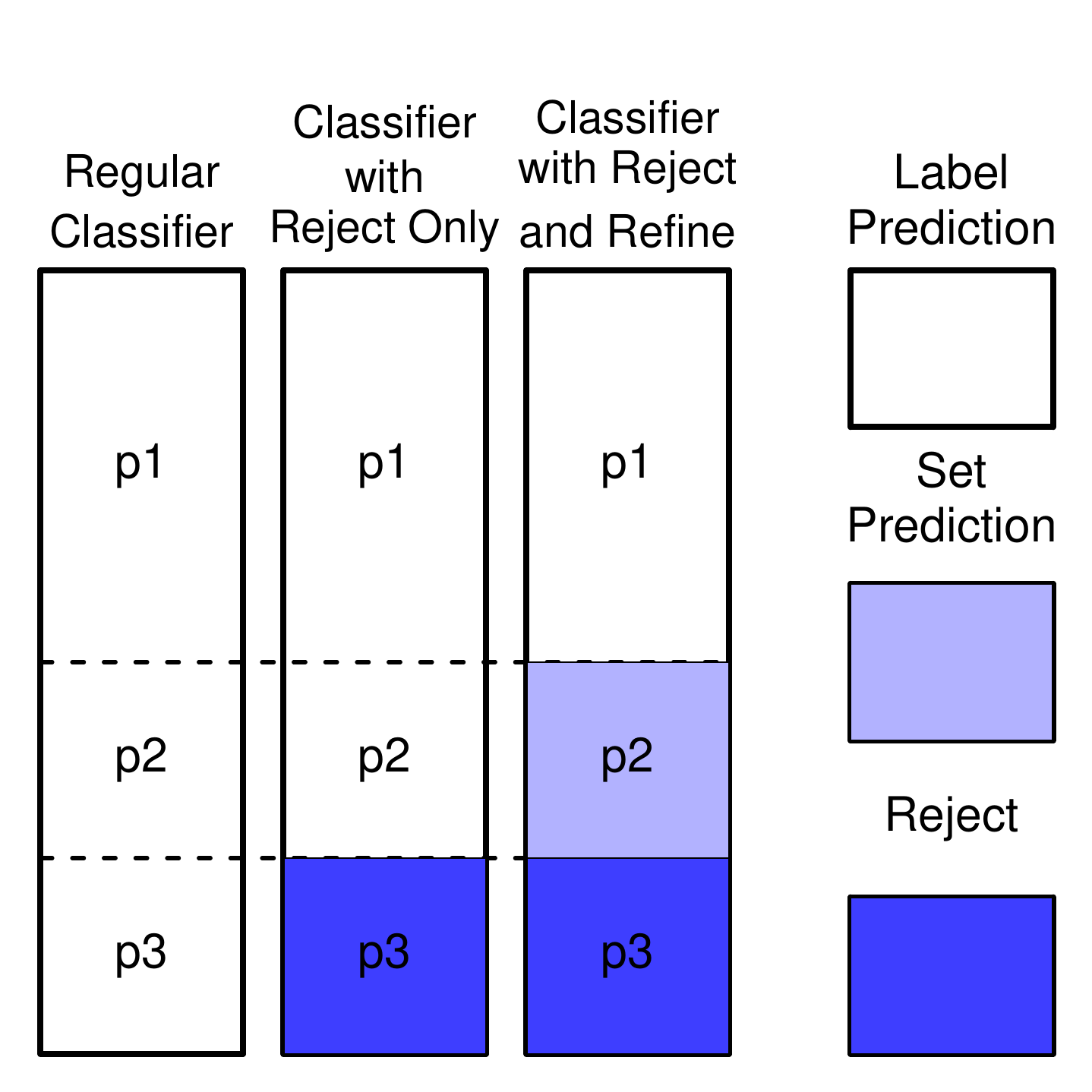}
\end{center}
\vspace{-2em}
\caption{{\small
Illustration of the partition of the test data based on prediction types.
}}
\vspace{-1em}
\label{simulationdemo}
\end{figure}

For all numerical problems in the current section, we study the performance of regular classifiers, classifiers with only the reject option, and classifiers with both reject and refine options. There are three possible prediction outcomes: (definite) label predictions, (refined) set predictions and rejections. Different types of predictions are shown using different colors in Figure~\ref{simulationdemo}. It can be seen that different classifiers have different capacity: regular classifiers can only provide label predictions while our classifiers with a refinement option can yield all three prediction types. We report classification performance on three disjoint subsets of observations, namely, $p1$, $p2$ and $p3$. The three subsets are defined as the observations which are label predicted, set predicted and rejected, respectively, by the classifier with both reject and refine options.

We report the misclassification error for each observation subset for each classifier. No misclassification rate is reported for rejected observations. For observations that are refined by our classifier ($p2$), we report the mis-refinement rate, which is defined as the proportion of observations whose true class labels are not in the prediction sets. We also report the empirical $0$-$d$-$1$ loss for the whole test data set for each classifier, where we count cost $1$ for each misclassification or mis-refinement, and cost $d$ for each rejection. The proportions of $p_1$, $p_2$ and $p_3$ are reported, since one may want to avoid large proportions of $p_2$ and $p_3$ unless necessary. We also report the proportion and mis-refinement rate for selected sets of class labels when they are of interest for the discussion.

For comparison purpose, we also use classifiers with probability estimation, and plug in the estimates into the Bayes rule (in Proposition \ref{prop1}) to achieve a reject option. The proportion of the label predicted and rejected observations by this approach are calculated, and the misclassification rate for the label prediction set is reported. The overall empirical $0$-$d$-$1$ loss is reported as well.

We conduct 100 replications for each example and report the average.

\subsection{Simulations}
\label{simulation}

We consider three simulated examples to assess the performance of the proposed methods. We focus on linear learning here and consider the Soft-LUM classifier loss \citep[Soft;][]{LUM}, the DWD loss, and the SVM loss. Each loss is associated with one regular classifier, one with rejection only and one with both reject and refine options. Moreover, we implement the probability estimation method associated with Theorem 3 of \citet{MAC}.

To select the best tuning parameters $s$ and $\delta$, we choose from a candidate set $\Lambda \times \Delta$ the best pair that minimizes the empirical $0$-$d$-$1$ loss on a separate tuning data set, where $\Lambda$ consists of $30$ $\lambda$ values, and $\Delta = \{0.3 ,0.25  ,\ldots,0.05 ,0\} \cdot \max_{i,j} |\langle \mY_j,\hat{\bdf}(\bx_i)\rangle|$. The multiplicative constant $\max_{i,j} |\langle \mY_j,\hat{\bdf}(\bx_i)\rangle|$ is used to scale for the magnitude of the angle margins. This is because when $p$ is large, a severe regularization $J(\bdf)$ is often needed, which would shrink the magnitude of $\hat{\bdf}$ \citep{refit}. In this case, using a fixed set of candidate values in $\Delta$ could be suboptimal. Note that letting $\delta=0$ shuts off reject and refine options. To illustrate the effect of $a$ on the reject and refinement results, we fit the classifiers with several $a$ values between $a_1$ and $a_2$, but show the results for the best one only to save some space. More details are included in the Supplementary Materials.

\begin{figure}[!htb]
\vspace{-2em}
\begin{center}
\subfigure[
]{\includegraphics[angle=0,width=0.3\textwidth,totalheight=0.33\textwidth]{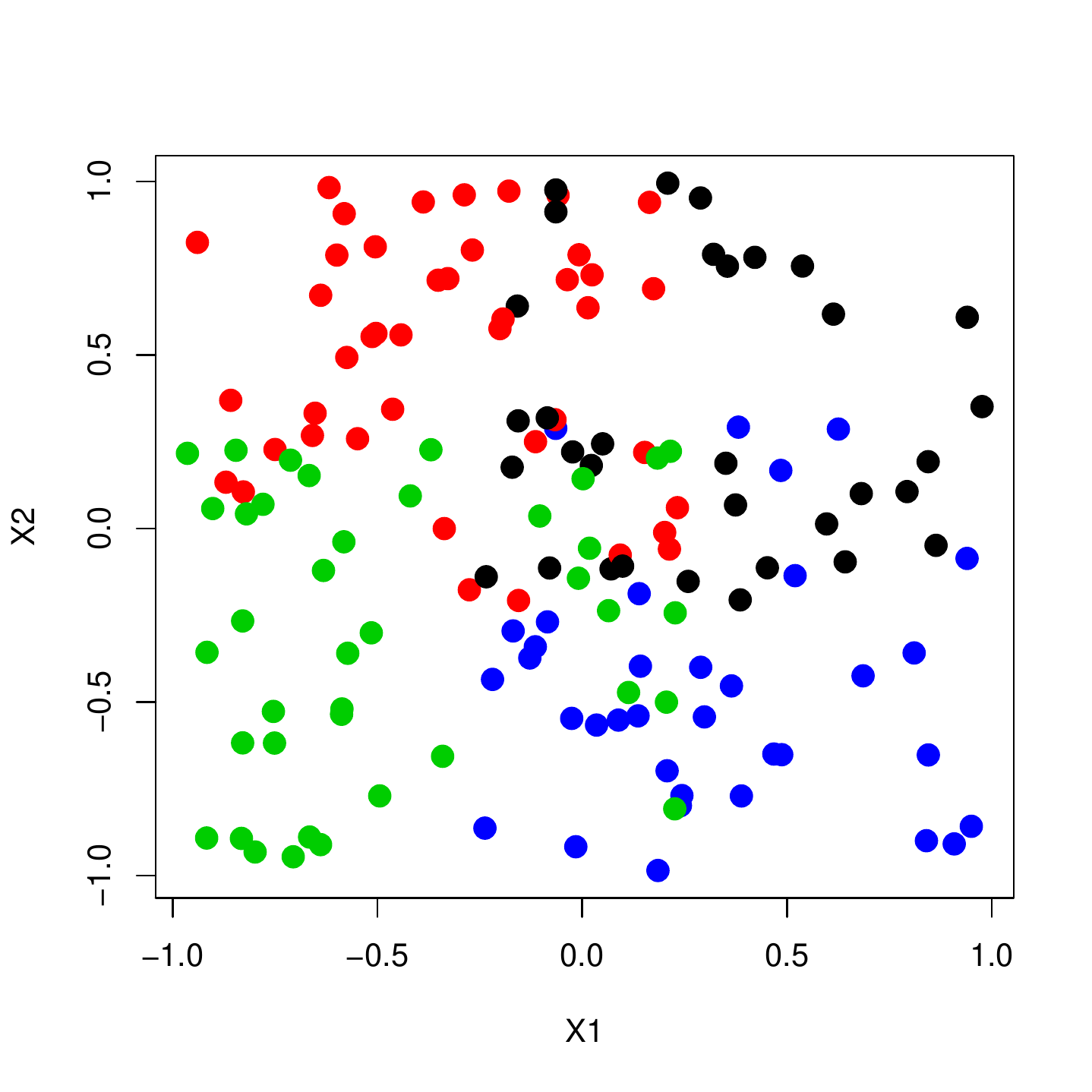}}
\subfigure[
]{\includegraphics[angle=0,width=0.3\textwidth,totalheight=0.33\textwidth]{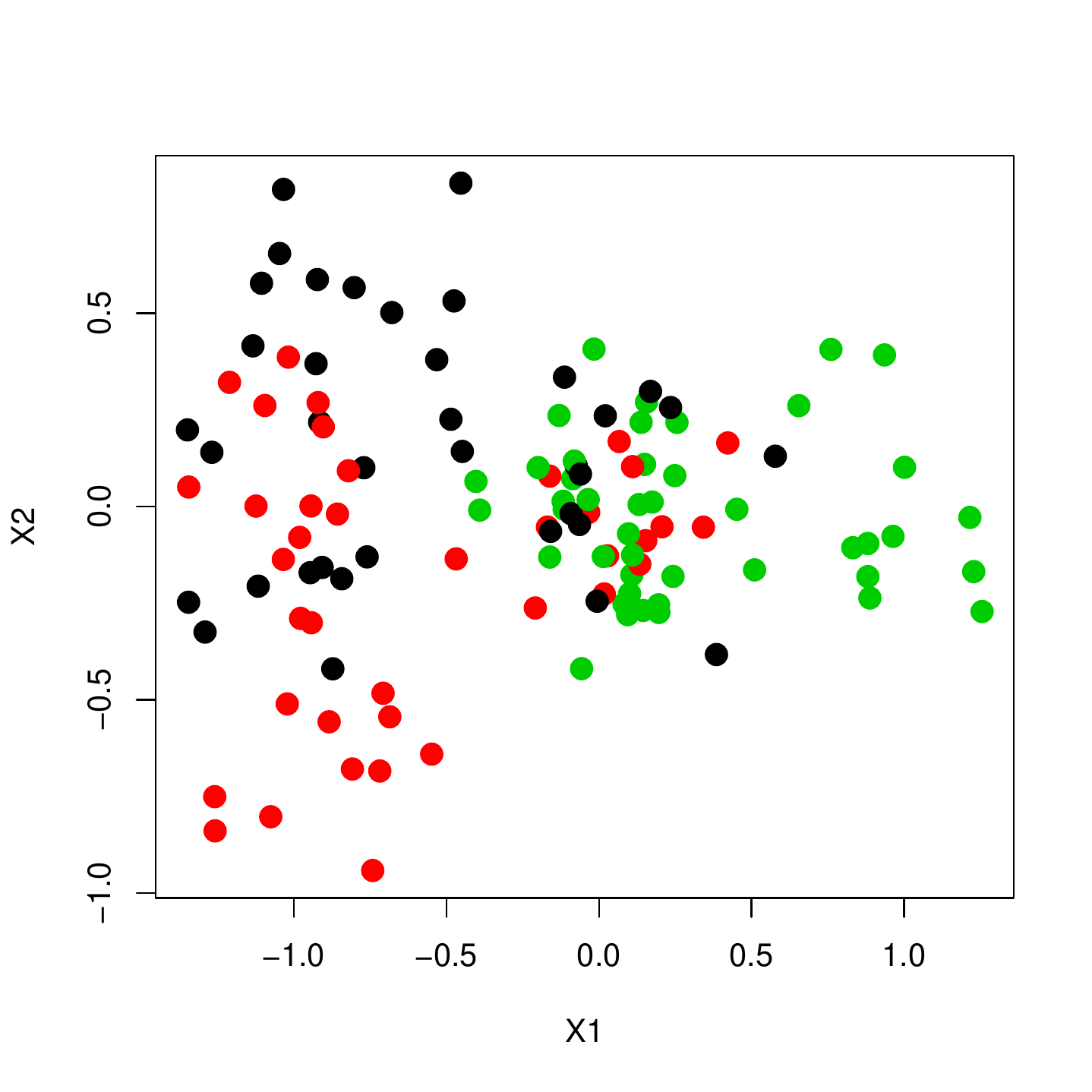}}
\subfigure[
]{\includegraphics[angle=0,width=0.3\textwidth,totalheight=0.33\textwidth]{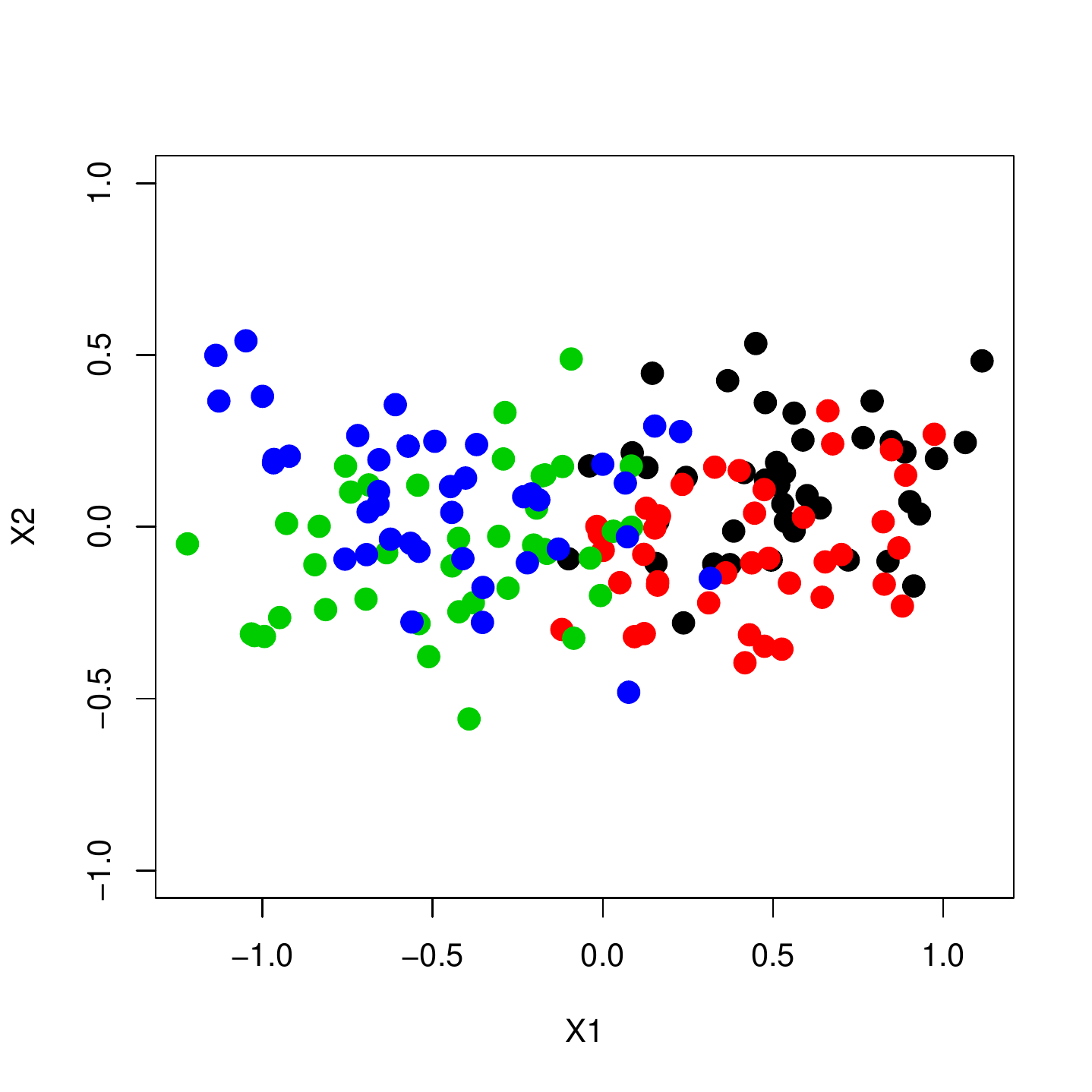}}
\end{center}
\vspace{-2em}
\caption{{\small
Plots of marginal distributions of $x_1$ and $x_2$ for Examples 1 (a), 2 (b), and 3 (c).
}}
\vspace{-1em}
\label{simufig}
\end{figure}

\noindent {\bf Example 1:} A four-class example with equal prior probabilities. We first generate two covariates that determine the true class distributions. In particular, $\bx \mid Y=j$, $j=1,2,3,4$, are uniformly distributed in $[-0.3,1] \times [-0.3, 1]$,  $[-0.3,1] \times [-1, 0.3]$, $[-1, 0.3] \times [-1, 0.3]$, and $[-1, 0.3] \times [-0.3,1]$ respectively. See the left panel of Figure~\ref{simufig} for a typical example on the first two dimensions. We then add 98 noise covariates. The training and tuning data sets are of size $150$ respectively, and the test data set is of size $12000$. In this example we let $d=0.6$, and report the behavior of the Soft loss using the $L_1$ penalty only.

\noindent {\bf Example 2:} A three-class example with equal prior probabilities. The true classification signal depends on two predictors, and the marginal distributions of $\bX \mid Y=j$, $j=1,2,3$, follow $\frac{1}{3} N\big( (-\sqrt{3}/2,1/2)^T, \sigma^2 I_2 \big) + \frac{1}{3} N \big( (-1,0)^T, \sigma^2 I_2 \big) + \frac{1}{3} N \big( (0, 0)^T, \sigma^2 I_2 \big),$  $\frac{1}{3} N\big( (-\sqrt{3}/2,-1/2)^T, \sigma^2 I_2 \big) + \frac{1}{3} N \big( (-1,0)^T, \sigma^2 I_2 \big) + \frac{1}{3} N \big( (0,0)^T, \sigma^2 I_2 \big),$ and
$\frac{2}{3} N\big( (1,0)^T, \sigma^2 I_2 \big) + \frac{1}{3} N \big( (0,0)^T, \sigma^2 I_2 \big)$ respectively where $\sigma=0.2$. We then add 398 noise covariates. The training and tuning sample sizes are both $120$, and the test sample size is $12000$. See the middle panel of Figure~\ref{simufig}. In this example we let $d=0.5$, and report the results for the SVM (hinge) loss using the $L_2$ penalty only. Many observations are confusable only between classes 1 and 2, hence the proportion and error rate for predictions $\{1,2\}$ are reported separately.

\noindent {\bf Example 3:} A four-class problem where the classification signal between classes 1 and 2, and that between classes 3 and 4, are confusable. We let the class label depend on only two predictors. In particular, for class $j$, the marginal distribution of $x_1$ and $x_2$ is normal with $\sigma = 0.2$, and the corresponding mean is uniformly distributed on the line segment between $(0,0)^T$ and $(z_1^j,z_2^j)^T; \ j=1,\ldots,4$, where $(z_1^1,z_2^1) = (1,0.2)$, $(z_1^2,z_2^2) = (1,-0.2)$, $(z_1^3,z_2^3) = (-1,0.2)$, and $(z_1^4,z_2^4) = (-1,-0.2)$. See the right panel of Figure~\ref{simufig}. We then add 98 noise covariates. The training and tuning data sets are both of size $160$, and the test data set size is $10000$. In this example, we choose $d=0.5$, and report the performance of DWD loss with the $L_1$ penalty. The proportions and error rates of prediction $\{1,2\}$ and $\{3,4\}$ are reported.

All the noise covariates added are $i.i.d.$ $N(0,0.01)$. To save space, we only report selected results here in Table \ref{simu_rep_tab1}, while more results can be found in the Supplementary Materials. We collect some key observations below.

\begin{table*}[!t]
{
\small\hfill
\begin{tabular}{ c | c | l | c | c | c || c | c  }
\hline\hline
\multicolumn{3}{c | }{Example 1, Soft with $a=a_2$} & Regular & Reject & R\&R & \multicolumn{2}{c }{Probability Method} \\
\hline
 & \multicolumn{2}{c | }{Proportion} & \multicolumn{3}{c || }{Error} & Proportion & Error \\
\hline
p1 & \multicolumn{2}{c | }{49.43} & 28.80 & 27.58 & 27.58 & \multirow{3}{*}{52.26} &  \multirow{3}{*}{30.17} \\
\cline{1-6}
\multirow{2}{*}{p2} & \multirow{2}{*}{28.97} & size 2: 24.62 & \multirow{2}{*}{45.89} & \multirow{2}{*}{45.35} & \multirow{2}{*}{1.581} & & \\
\cline{3-3}
    & & size 3: 4.349 & & & & & \\
\hline
p3 & \multicolumn{2}{c | }{21.60} & 69.61 & - & - & 47.74 & - \\
\hline
Overall & \multicolumn{2}{c | }{100.0} & 41.92 & 39.32 & 27.47 & 100.0 & 45.64 \\
\hline\hline\hline
\multicolumn{3}{c | }{Example 2, SVM with $a=a_1$} & Regular & Reject & R\&R & \multicolumn{2}{c }{Probability Method} \\
\hline
 & \multicolumn{2}{c | }{Proportion} & \multicolumn{3}{c || }{Error} & Proportion & Error \\
\hline
p1 & \multicolumn{2}{c | }{52.53} & 28.15 & 27.81 & 27.81 & \multirow{3}{*}{42.40} &  \multirow{3}{*}{25.85} \\
\cline{1-6}
\multirow{2}{*}{p2} & \multicolumn{2}{l | }{size2: 12.90} & \multirow{2}{*}{48.57} & \multirow{2}{*}{51.00} & \multirow{2}{*}{11.26} & & \\
& \multicolumn{2}{c | }{$\quad\lfloor \ \{1,2\}$: 73.1\%} & & & & & \\
\hline
p3 & \multicolumn{2}{c | }{34.57} & 53.01 & - & - & 57.60 & - \\
\hline
Overall & \multicolumn{2}{c | }{100.0} & 39.57 & 38.91 & 33.33 & 100.0 & 40.66 \\
\hline\hline\hline
\multicolumn{3}{c | }{Example 3, DWD with $a=a_2$} & Regular & Reject & R\&R & \multicolumn{2}{c }{Probability Method} \\
\hline
 & \multicolumn{2}{c | }{Proportion} & \multicolumn{3}{c || }{Error} & Proportion & Error \\
\hline
p1 & \multicolumn{2}{c | }{45.58} & 25.68 & 26.11 & 26.11 & \multirow{5}{*}{36.97} &  \multirow{5}{*}{24.72} \\
\cline{1-6}
\multirow{4}{*}{p2} & \multirow{4}{*}{ 23.26 } & size 2: 19.71 & \multirow{4}{*}{36.45} & \multirow{4}{*}{35.71} & \multirow{4}{*}{1.771} & & \\
    & & $\quad\lfloor \ \{1,2\}$: 40.3\% & & & & & \\
    & & $\quad\lfloor \ \{3,4\}$: 42.9\% & & & & & \\
    \cline{3-3}
    & & size 3: 3.549 & & & & & \\
\hline
p3 & \multicolumn{2}{c | }{31.16} & 56.02 & - & - & 63.03 & - \\
\hline
Overall & \multicolumn{2}{c | }{100.0} & 36.44 & 35.97 & 27.90 & 100.0 & 41.78 \\
\hline\hline
\end{tabular}
}
\hfill{} \caption{Simulation results for Examples 1, 2 and 3, with the Soft loss, the SVM loss and the DWD loss respectively. The average proportion and misclassification or mis-refinement rate on each observation set $p_1$, $p_2$ or $p_3$ for each of the three classifiers (regular, with rejection only, and with both reject and refine options) are reported over 100 replications. A probability estimation method for reject option is also compared. The overall empirical $0$-$d$-$1$ loss is reported for each classifier and each example. Proportions for selected prediction sets are reported as well. The results show improved overall loss by our methods and successful identification of a subset of most confusing classes.}
\label{simu_rep_tab1}
\end{table*}

\noindent$\bullet$~ For the probability estimation plug-in method, its overall error is greater than our proposed methods, and can be even greater than that of the regular classifier without a reject option. This is because for high-dimensional problems, accurate probability estimation is too difficult, leading to degenerated performance for classification.

\noindent$\bullet$~ Our proposed classifiers with the rejection option can lead to less overall error, compared to traditional classifiers which only provide label predictions. The large values of the errors on $p_3$ for the regular classifiers (all of which are greater than 50\%),  indicates that the reject option is able to identify the set of testing observations which are most difficult to be classified.

\noindent$\bullet$~ The refine option can provide very accurate set prediction, with very low mis-refinement rates. The usefulness of the classifiers with both the reject and refine options is also reflected by the decreased overall error rate.
\begin{enumerate}
		\item Compared to the classifier with only the reject option, the classifier with both reject and refine options can avoid misclassification for subset $p_2$: this can be seen from the reduction from misclassification rate 45.35\% (51.00\%, 35.71\%, resp.) to mis-refinement rate 1.581\% (11.26\%, 1.771\%, resp.)
		\item Another advantage of the refine option is that it can identify class labels that are most confusing to each other. In Example 2, 73.1\% of the size 2 set predictions are $\{1,2\}$, while class 1 and class 2 indeed exist a two-way confusion. In Example 3, about 40\% of the size 2 set predictions are $\{1,2\}$ (and another 40\% for $\{3,4\}$.) Hence, a researcher can conclude that the intrinsic difference between classes $1$ and $2$, or $3$ and $4$, is relatively small. In genetic research, this information can be used to verify that two diseases are similar, or can be used to introduce new studies on the corresponding causations.
\end{enumerate}

\noindent$\bullet$~ As the parameter $a$ grows (not shown here for the sake of space), our proposed methods become more conservative. In particular, the proportion of rejected observations increases as $a$ increases. On the other hand, the effect of $a$ on the classification performance changes under various settings, and there is no single $a$ that works uniformly the best for all problems. Our numerical experience shows that the best $a$ for a given problem is often at either $a_1$ or $a_2$ (defined in Proposition~\ref{upperandlowerfora}.) Hence, for real applications, we recommend to train the classifier with $a=a_1$ and $a=a_2$, and select the one with the better performance.

\subsection{Real Data Analysis}
\label{realdata}

In this section, we illustrate the use of our methods for the Glioblastoma Multiforme Cancer data set \citep[GBM,][]{cancercell} and the normalized handwritten digits data set scanned from envelopes by the U.S. Postal Service \citep[ZIP,][]{Hastie2009}.

In the GBM data set, there are 4 subtypes of Glioblastoma Multiforme cancer, namely, Classical, Mesenchymal, Neural and Proneural, and within each type we have 92, 111, 56, 97 patients, respectively. The gene expression levels on 16548 genes are measured as predictors to characterize the cancer subtypes. We normalize the data set so that each predictor has mean 0 and sample variance 1. As a demonstration, we use $d=0.4$, the Soft loss, and the $L_2$ penalty. To select the best tuning parameters, we split the data set into 6 groups of observations whose sizes are roughly the same, choose one group as the test data set, and perform 5-fold cross validations on the remaining observations. We report the average result over 100 random splits. To alleviate the computational burden, we choose 2000 genes with the greatest median absolute deviation values based on the training sample for each split.

\begin{table*}[!t]
{
\small \hfill{}
\begin{tabular}{ c | c | l | c | c | c || c | c  }
\hline\hline
\multicolumn{3}{c | }{GBM, Soft with $a=a_1$} & Regular & Reject & R\&R & \multicolumn{2}{c }{Probability Method} \\
\hline
 & \multicolumn{2}{c | }{Proportion} & \multicolumn{3}{c || }{Error} & Proportion & Error \\
\hline
p1 & \multicolumn{2}{c | }{72.15} & 13.69 & 13.69 & 13.69 & \multirow{5}{*}{72.58} &  \multirow{5}{*}{13.78} \\
\cline{1-6}
\multirow{4}{*}{p2} & \multirow{4}{*}{18.99} & size 2: 17.14 & \multirow{4}{*}{41.35} & \multirow{4}{*}{39.53} & \multirow{4}{*}{3.724} & & \\
    & & $\quad\lfloor \ \{C,M\}$: 44.3\% & & & & & \\
    & & $\quad\lfloor \ \{N,P\}$: 33.7\% & & & & & \\
    \cline{3-3}
    & & size 3: 1.853 & & & & & \\
\hline
p3 & \multicolumn{2}{c | }{8.857} & 43.33 & - & - & 27.42 & - \\
\hline
Overall & \multicolumn{2}{c | }{100.0} & 21.85 & 20.97 & 14.13 & 100.0 & 21.25 \\
\hline\hline\hline
\multicolumn{3}{c | }{ZIP, DWD with $a=a_1$} & Regular & Reject & R\&R & \multicolumn{2}{c }{Probability Method} \\
\hline
 & \multicolumn{2}{c | }{Proportion} & \multicolumn{3}{c || }{Error} & Proportion & Error \\
\hline
p1 & \multicolumn{2}{c | }{97.05} & 2.087 & 2.087 & 2.087 & \multirow{3}{*}{98.90} &  \multirow{3}{*}{2.607} \\
\cline{1-6}
\multirow{2}{*}{p2} & \multicolumn{2}{l | }{size 2: 2.578} & \multirow{2}{*}{35.71} & \multirow{2}{*}{28.57} & \multirow{2}{*}{0.110} & & \\
& \multicolumn{2}{c | }{$\quad\lfloor \ \{4,9\}$: 55.9\%} & & & & & \\
\hline
p3 & \multicolumn{2}{c | }{0.368} & 95.14 & - & - & 1.104 & - \\
\hline
Overall & \multicolumn{2}{c | }{100.0} & 3.314 & 2.909 & 2.175 & 100.0 & 3.020 \\
\hline\hline
\end{tabular}}
\hfill{} \caption{Summary of analysis for the GBM data set (ZIP data set, resp.) with the Soft loss (the DWD loss, resp.) and $a=a_1$. The average proportion and misclassification or mis-refinement rate on each observation set $p_1$, $p_2$ or $p_3$ for each of the regular classifier, classifier with rejection only and classifier with both reject and refine options are reported over 100 splits. A probability method for reject option is also compared. The overall empirical $0$-$d$-$1$ loss is reported for each classifier and each example. Proportions for selected prediction sets are reported as well. In the GBM example, $\{C,M\}=\{\textrm{Classical},\textrm{Mesenchymal}\}$ and $\{N,P\}=\{\textrm{Neural}, \textrm{Proneural}\}$. The results show improved overall loss by our methods and successful identification of a subset of most confusing classes. }
\label{GBM_rep_tab}
\end{table*}

We include selected results for the GBM data set on the top half of Table~\ref{GBM_rep_tab}. More results can be found in the Supplementary Materials. Our proposed reject and refine options can often abstain from making label predictions on observations on which the classification signal is weak, which leads to a reduced overall loss. The mis-refinement rate on the refined observations ($p_2$) is very small compared to the label prediction error rates by the regular classifier and the reject only classifier. More interestingly, most of the set predictions occur for either $\{\textrm{Classical},\textrm{Mesenchymal}\}$ or $\{\textrm{Neural},\textrm{Proneural}\}$. This suggests that the GBM subtypes Classical and Mesenchymal, or Neural and Proneural, share some common characteristics in their genotypes. This finding is  consistent with both The Cancer Genome Atlas core samples or validation samples in \citet{cancercell}.

\begin{figure}[!t]
\begin{center}
\includegraphics[angle=0,width=0.9\textwidth,totalheight=0.35\textwidth]{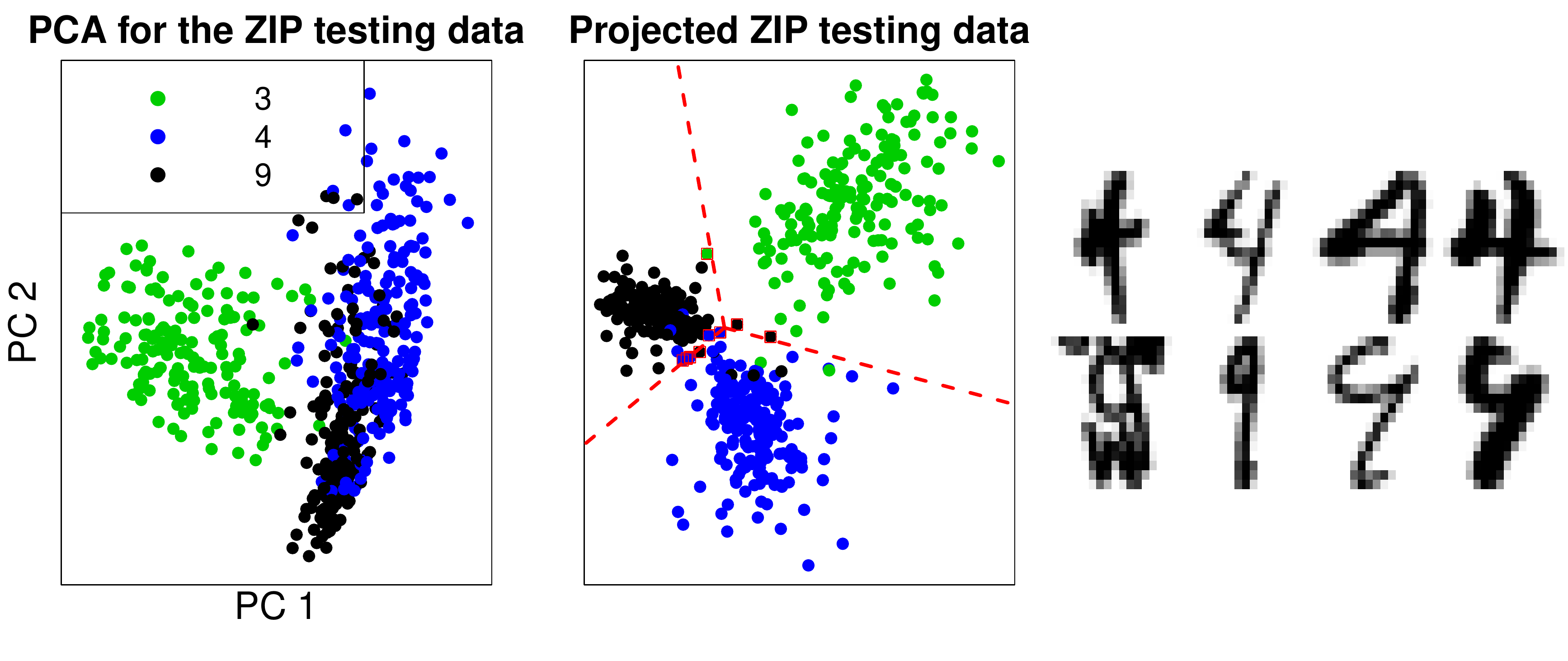}
\end{center}
\vspace{-2em}
\caption{{\small
Left: the PCA scatter plot. Middle: the test data mapped to $\mbbR^2$ using $\hat \bdf(\bx)\in\R^2$ in a typical split, where the dashed lines correspond to the classification boundaries, and observations with reject or refine prediction are identified as red squares. Right: some observations that often ($>80\%$ within the 100 splits) have refined prediction $\{4,9\}$.
}}
\label{ZIPfig}
\end{figure}

The ZIP data set has been extensively studied by many previous works. We choose categories ``3", ``4" and ``9" to demonstrate the effect of the refine option. For handwritten digits, it is sometimes difficult for machines to classify between ``4" and ``9", while the difference between ``3" and ``4" or ``3" and ``9" is more obvious. For visualization, we draw a PCA plot for the test data on the left panel of Figure~\ref{ZIPfig}. In the middle panel, we provide a scatter plot by projecting the sample to the 2D space using $\hat\bdf(\bx)\in\R^2$. In particular, observations with reject or refined set predications are shown in red squares. It can be seen that the observations which are refined are precisely those sitting on 2-way classification boundaries (shown as the dashed red lines), while most of them are between ``4'' and ``9''. In the analysis, we use $d=0.4$, the DWD loss, and the $L_2$ penalty. We normalize the data set before the analysis. To select the best tuning parameters, we split the training data set into two groups, and use one to train the classifier and the other for tuning. We report the average results of 100 splits.

The results for the ZIP data set are reported in the second half of Table~\ref{GBM_rep_tab}, while more comprehensive results can be found in the Supplementary Materials. Note that although there are only a few rejected observations ($<0.368\%$ on average), their misclassification rate is as high as 95.14\%, if not rejected. This stunningly high error rate justifies our reject option. Though there are only 2.578\% observations that are refined, the mis-refinement rate is as low as 0.110\%, almost always correct. The middle panel of Figure \ref{GBM_rep_tab} also suggests that the refinement decision is well deserved since the refined data points are in close vicinity to the classification boundaries. Lastly, it can be seen that, for quite a few observations, the classification signal is very vague between ``4" and ``9", which is consistent with our common sense (see the middle and right panels of Figure~\ref{ZIPfig}).

\section{Conclusion}\label{sec:conclude}

In this article, we enrich regular multicategory classification methods with reject and refine options. While a reject option adds to the capacity of an existing multicategory classifier, a refine option has the potential to open a new direction. Usually, statistical learning researches have been aiming to create an ``ultimate'' machine with perfect predictive power. However, sometimes the nature of the data or the data collection process has determined that a significant misclassification is inevitable for some observations. Many methods have been proposed but the obtained improvement is somewhat limited. When the cost of misclassification is too large to bear, it may be wise to take a step back and start to think of new ideas out of the box. A refine option may have opened a door to these. With a refine option, one can often successfully identify observations with a subset of most confusable class labels. Future resources can be allocated to these observations with a refined set of labels to acquire new data with better quality.

The aim of this article is to establish a new framework for classification. Many different loss functions and penalty functions may be incorporated into this framework for the purpose of reject and refine options. We provide a novel statistical learning theory, with emphasis on diverging dimensions and diverging numbers of classes. Future work will be devoted to how to better utilize the refined set predictions. Many new research topics will follow this new learning tool.

\section*{Appendix}
In this appendix, we provide brief outlines of the proofs of Theorems~\ref{excessellriskthm}, \ref{excessellriskthm_kernel} and \ref{fast_rate_thm}.

\noindent{\bf Theorem \ref{excessellriskthm} and Theorem \ref{excessellriskthm_kernel}}\\
There are two major steps in the proof of Theorems~\ref{excessellriskthm} and~\ref{excessellriskthm_kernel}. The first step is to decompose the excess $\ell$-risk into the estimation error and approximation error. Then we show that the probability of the estimation error exceeding $O(s k r \log(r^{-1}))$ for the $L_1$ penalty, or $O(\sqrt{ps} k r \log(r^{-1}))$ for the $L_2$ penalty, can be written in terms of a concentration inequality indexed by a scaled empirical process. The second step is to obtain a suitable probability upper bound of this concentration inequality. To this end, one can use the chaining technique, which discretizes the functional space of the optimization problem, hence decomposing the corresponding probability into several parts. For each part, the probability can be controlled by established concentration inequalities. See Theorem A.2 in \citet{Wang2007} for an example.

Therefore, the question boils down to control the complexity of the discretized functional space. A common approach to depict such complexity in the literature is to use the entropy numbers. In the Supplementary Materials, for linear and kernel learning, we introduce Lemmas 2 and 4 respectively, to control the complexity of the corresponding functional spaces for the empirical processes, in terms of their $L_2$ entropy numbers. In particular, we show that for a small and positive $\epsilon$, the $\epsilon$-entropy numbers for linear and kernel learning are in the order of $O(\epsilon^{-2})$ under mild conditions. Consequently, we can prove the desired concentration inequality.

It should be noted that, although the orders of the entropy numbers for linear and kernel learning are similar, the techniques used are quite different. In particular, in linear learning, we treat the functional space as a convex hull of $2p$ functions, which leads to a bound on the entropy number. For kernel learning, we consider the natural embedding of the kernel function into the regular $L_2$ functional space consisting of continuous functions on the domain of $\bx$. Such embedding can be shown to be absolutely 2-summing with 2-summing norm no larger than 1. Hence we can bound the entropy number of this embedding operator (which can be shown to be the same as the entropy number of the original kernel space) by its corresponding approximation numbers, which can be further bounded by Carl's inequality between approximation and entropy numbers.

\noindent{\bf Theorem \ref{fast_rate_thm}}\\
Theorem~\ref{fast_rate_thm} extends the well established results on fast rate of convergence from binary classifiers to multicategory ones. The key to the proof is to find a pseudo-norm that can be used to both upper and lower bound the conditional excess $\ell$-risk $g_{\bdf}(\bx,y) = \sum_{j \ne y} \ell\{\langle \bdf,\mY_j \rangle\} - \sum_{j \ne y} \ell\{\langle \bdf^*,\mY_j \rangle\}$ (up to constants). In \citet{bartlett2008classification}, as the modified hinge loss function $\psi(u)$ is piecewise linear, and remains flat for large $u$, one can use $\rho(f_1,f_2) \propto |f_1-f_2|$ as the pseudo-norm. However, for more general loss functions, especially differentiable loss functions, an $L_1$ type pseudo-norm cannot lower bound the conditional excess $\ell$-risk. Therefore, we employ the (squared) $L_2$ type pseudo-norm in this proof. With the low noise assumption, we can show that the class $\{g_{\bdf}(\bx,y)\}$ is a Bernstein class with the Bernstein exponent $\alpha/(1+\alpha)$. The next step is to apply the symmetrization technique, and show that the estimation error can be (up to a constant) bounded by a tail probability plus a small term that converges to zero at a very fast speed, where the tail probability term is indexed by an empirical process of $\{g_{\bdf}(\bx,y)\}$. At this stage, we can employ Bernstein's inequality to bound the corresponding tail probability. As $\{g_{\bdf}(\bx,y)\}$ is a Bernstein class, the variance term in the power of the upper bound in Bernstein's inequality can be bounded by a linear term of $\E g_{\bdf}(\bx,y)$. Combined with an upper bound on the entropy number for Gaussian kernel space, we can prove the desired result in Theorem~\ref{fast_rate_thm}.

\section*{Supplementary Materials}

\begin{description}
	\item[SM1:] Detailed proofs of Proposition \ref{proposition2}, Proposition \ref{upperandlowerfora}, Theorem \ref{excessellriskthm}, Theorem \ref{excessellriskthm_kernel} and Theorem \ref{fast_rate_thm}; derivations of the implementations; extended numerical results. (.pdf file)
\end{description}

\bibliographystyle{asa}
\bibliography{ref_reject}

\begin{thebibliography}{52}
\newcommand{\enquote}[1]{``#1''}
\expandafter\ifx\csname natexlab\endcsname\relax\def\natexlab#1{#1}\fi

\bibitem[{Bartlett et~al.(2006)Bartlett, Jordan, and McAuliffe}]{Bartlett06}
Bartlett, P.~L., Jordan, M.~I., and McAuliffe, J.~D. (2006),
  \enquote{Convexity, {C}lassification, and {R}isk {B}ounds,} \textit{Journal
  of the American Statistical Association}, 101, 138--156.

\bibitem[{Bartlett and Wegkamp(2008)}]{bartlett2008classification}
Bartlett, P.~L. and Wegkamp, M.~H. (2008), \enquote{{Classification with a
  Reject Option Using a Hinge Loss},} \textit{Journal of Machine Learning
  Research}, 9, 1823--1840.

\bibitem[{Blanchard et~al.(2008)Blanchard, Bousquet, and
  Massart}]{Blanchard2008}
Blanchard, G., Bousquet, O., and Massart, P. (2008), \enquote{{Statistical
  Performance of Support Vector Machines},} \textit{Annals of Statistics}, 36,
  489--531.

\bibitem[{Bousquet and Elisseeff(2002)}]{bousquet2002stability}
Bousquet, O. and Elisseeff, A. (2002), \enquote{{Stability and
  Generalization},} \textit{Journal of Machine Learning Research}, 2, 499--526.

\bibitem[{Boyd et~al.(2011)Boyd, Parikh, Chu, Peleato, and
  Eckstein}]{boyd2011distributed}
Boyd, S., Parikh, N., Chu, E., Peleato, B., and Eckstein, J. (2011),
  \enquote{{Distributed Optimization and Statistical Learning via the
  Alternating Direction Method of Multipliers},} \textit{Foundations and Trends
  in Machine Learning}, 3, 1--122.

\bibitem[{Cai et~al.(2014)Cai, Liu, and Zhou}]{cai2012estimating}
Cai, T.~T., Liu, W., and Zhou, H.~H. (2014), \enquote{{Estimating Sparse
  Precision Matrix: Optimal Rates of Convergence and Adaptive Estimation},}
  \textit{Annals of Statistics}, forthcoming.

\bibitem[{Chen et~al.(2004)Chen, Wu, Ying, and Zhou}]{chen2004support}
Chen, D.-R., Wu, Q., Ying, Y., and Zhou, D.-X. (2004), \enquote{{Support Vector
  Machine Soft Margin Classifiers: Error Analysis},} \textit{Journal of Machine
  Learning Research}, 5, 1143--1175.

\bibitem[{Chow(1970)}]{chow1970optimum}
Chow, C. (1970), \enquote{{On Optimum Recognition Error and Reject Tradeoff},}
  \textit{Information Theory, IEEE Transactions on}, 16, 41--46.

\bibitem[{Cortes and Vapnik(1995)}]{Cortes95}
Cortes, C. and Vapnik, V.~N. (1995), \enquote{{Support Vector Networks},}
  \textit{Machine Learning}, 20, 273--297.

\bibitem[{Crammer and Singer(2001)}]{Crammer01}
Crammer, K. and Singer, Y. (2001), \enquote{On the {A}lgorithmic
  {I}mplementation of {M}ulticlass {K}ernel-based {V}ector {M}achines,}
  \textit{Journal of Machine Learning Research}, 2, 265--292.

\bibitem[{Donoho(1995)}]{Donoho1995De}
Donoho, D.~L. (1995), \enquote{{De-noising by soft-thresholding},}
  \textit{Information Theory, IEEE Transactions on}, 41, 613--627.

\bibitem[{El-Yaniv and Wiener(2010)}]{el2010foundations}
El-Yaniv, R. and Wiener, Y. (2010), \enquote{{On the Foundations of Noise-free
  Selective Classification},} \textit{Journal of Machine Learning Research},
  11, 1605--1641.

\bibitem[{Fan and L\"u(2008)}]{fan2008sure}
Fan, J. and L\"u, J. (2008), \enquote{{Sure Independence Screening for
  Ultrahigh Dimensional Feature Space},} \textit{Journal of the Royal
  Statistical Society: Series B}, 70, 849--911.

\bibitem[{Freund and Schapire(1997)}]{Freund97}
Freund, Y. and Schapire, R.~E. (1997), \enquote{A {D}esicion-theoretic
  {G}eneralization of {O}n-line {L}earning and an {A}pplication to {B}oosting,}
  \textit{Journal of Computer and System Sciences}, 55, 119--139.

\bibitem[{Friedman et~al.(2010)Friedman, Hastie, and Tibshirani}]{Hastie10}
Friedman, J.~H., Hastie, T.~J., and Tibshirani, R.~J. (2010),
  \enquote{Regularization {P}aths for {G}eneralized {L}inear {M}odels via
  {C}oordinate {D}escent,} \textit{Journal of Statistical Software}, 33, 1--22.

\bibitem[{Fumera and Roli(2002)}]{fumera2002support}
Fumera, G. and Roli, F. (2002), \enquote{{Support Vector Machines with Embedded
  Reject Option},} in \textit{Pattern Recognition with Support Vector
  Machines}, Springer, pp. 68--82.

\bibitem[{Fumera et~al.(2000)Fumera, Roli, and Giacinto}]{fumera2000reject}
Fumera, G., Roli, F., and Giacinto, G. (2000), \enquote{{Reject Option with
  Multiple Thresholds},} \textit{Pattern recognition}, 33, 2099--2101.

\bibitem[{F\"urnkranz and H\"ullermeier(2010)}]{Fuernkranz2010Preference}
F\"urnkranz, J. and H\"ullermeier, E. (2010), \enquote{Preference Learning,} in
  \textit{Encyclopedia of Machine Learning}, eds. Sammut, C. and Webb, G.,
  Springer US, pp. 789--795.

\bibitem[{Gupta et~al.(2014)Gupta, Bengio, and Weston}]{gupta2014training}
Gupta, M.~R., Bengio, S., and Weston, J. (2014), \enquote{{Training Highly
  Multiclass Classifiers},} \textit{Journal of Machine Learning Research}, 15,
  1461--1492.

\bibitem[{Hastie et~al.(2009)Hastie, Tibshirani, and Friedman}]{Hastie2009}
Hastie, T.~J., Tibshirani, R.~J., and Friedman, J.~H. (2009), \textit{The
  Elements of Statistical Learning}, New York: Springer, 2nd ed.

\bibitem[{Herbei and Wegkamp(2006)}]{herbei2006classification}
Herbei, R. and Wegkamp, M.~H. (2006), \enquote{{Classification with Reject
  Option},} \textit{Canadian Journal of Statistics}, 34, 709--721.

\bibitem[{Le~Capitaine and Fr{\'e}licot(2010)}]{le2010optimum}
Le~Capitaine, H. and Fr{\'e}licot, C. (2010), \enquote{{An Optimum
  Class-rejective Decision Rule and its Evaluation},} in \textit{Pattern
  Recognition (ICPR), 2010 20th International Conference on}, IEEE, pp.
  3312--3315.

\bibitem[{Lee et~al.(2004)Lee, Lin, and Wahba}]{Lee04}
Lee, Y., Lin, Y., and Wahba, G. (2004), \enquote{Multicategory {S}upport
  {V}ector {M}achines, {T}heory, and {A}pplication to the {C}lassification of
  {M}icroarray {D}ata and {S}atellite {R}adiance {D}ata,} \textit{Journal of
  the American Statistical Association}, 99, 67--81.

\bibitem[{Liu and Shen(2006)}]{multipsi06}
Liu, Y. and Shen, X. (2006), \enquote{Multicategory $\psi$-learning,}
  \textit{Journal of the American Statistical Association}, 101, 500--509.

\bibitem[{Liu and Yuan(2011)}]{RMSVM}
Liu, Y. and Yuan, M. (2011), \enquote{Reinforced {M}ulticategory {S}upport
  {V}ector {M}achines,} \textit{Journal of Computational and Graphical
  Statistics}, 20, 901--919.

\bibitem[{Liu et~al.(2011)Liu, Zhang, and Wu}]{LUM}
Liu, Y., Zhang, H.~H., and Wu, Y. (2011), \enquote{Soft or {H}ard
  {C}lassification? {L}arge {M}argin {U}nified {M}achines,} \textit{Journal of
  the American Statistical Association}, 106, 166--177.

\bibitem[{Mai and Zou(2012)}]{mai2012kolmogorov}
Mai, Q. and Zou, H. (2012), \enquote{{The Kolmogorov Filter for Variable
  Screening in High-dimensional Binary Classification},} \textit{Biometrika},
  100, 229--234.

\bibitem[{Marron et~al.(2007)Marron, Todd, and Ahn}]{Marron07}
Marron, J.~S., Todd, M., and Ahn, J. (2007), \enquote{{Distance Weighted
  Discrimination},} \textit{Journal of the American Statistical Association},
  102, 1267--1271.

\bibitem[{Qiao and Zhang(2015)}]{qiao2015flexible}
Qiao, X. and Zhang, L. (2015), \enquote{{Flexible High-dimensional
  Classification Machines and Their Asymptotic Properties},} \textit{Journal of
  Machine Learning Research}, forthcoming (arXiv preprint:1310.3004).

\bibitem[{Sch\"olkopf and Smola(2002)}]{Schlkopf2002}
Sch\"olkopf, B. and Smola, A.~J. (2002), \textit{Learning with Kernels: Support
  Vector Machines, Regularization, Optimization, and Beyond (Adaptive
  Computation and Machine Learning)}, The MIT Press.

\bibitem[{Shawe-Taylor and Cristianini(2004)}]{kernelbook}
Shawe-Taylor, J.~S. and Cristianini, N. (2004), \textit{Kernel Methods for
  Pattern Analysis}, Cambridge University Press, 1st ed.

\bibitem[{Shen et~al.(2003)Shen, Tseng, Zhang, and Wong}]{Shen03}
Shen, X., Tseng, G.~C., Zhang, X., and Wong, W.~H. (2003), \enquote{On
  $\psi$-learning,} \textit{Journal of the American Statistical Association},
  98, 724--734.

\bibitem[{Steinwart and Christmann(2008)}]{steinwart2008support}
Steinwart, I. and Christmann, A. (2008), \textit{Support Vector Machines},
  Springer.

\bibitem[{Steinwart and Scovel(2007)}]{Ingo07}
Steinwart, I. and Scovel, C. (2007), \enquote{{Fast Rates for Support Vector
  Machines using Gaussian Kernels},} \textit{Annals of Statistics}, 35,
  575--607.

\bibitem[{Tax and Duin(2008)}]{tax2008growing}
Tax, D. M.~J. and Duin, R. P.~W. (2008), \enquote{{Growing a Multi-class
  Classifier with a Reject Option},} \textit{Pattern Recognition Letters}, 29,
  1565--1570.

\bibitem[{Tsybakov(2004)}]{tsybakov2004optimal}
Tsybakov, A.~B. (2004), \enquote{{Optimal Aggregation of Classifiers in
  Statistical Learning},} \textit{Annals of Statistics}, 135--166.

\bibitem[{van~der Vaart and Wellner(2000)}]{EPbook}
van~der Vaart, A.~W. and Wellner, J.~A. (2000), \textit{Weak {C}onvergence and
  {E}mpirical {P}rocesses with {A}pplication to {S}tatistics}, Springer, 1st
  ed.

\bibitem[{Vapnik(1998)}]{Vapnik98}
Vapnik, V.~N. (1998), \textit{Statistical Learning Theory}, New York: Wiley.

\bibitem[{Verhaak et~al.(2010)Verhaak, Hoadley, Purdom, Wang, Qi, Wilkerson,
  Miller, Ding, Golub, Mesirov, Alexe, Lawrence, O'Kelly, Tamayo, Weir,
  Gabriel, Winckler, Gupta, Jakkula, Feiler, Hodgson, James, Sarkaria, Brennan,
  Kahn, Spellman, Wilson, Speed, Gray, Meyerson, Getz, Perou, Hayes, and
  {Cancer Genome Atlas Research Network}}]{cancercell}
Verhaak, R.~G., Hoadley, K.~A., Purdom, E., Wang, V., Qi, Y., Wilkerson, M.~D.,
  Miller, C.~R., Ding, L., Golub, T., Mesirov, J.~P., Alexe, G., Lawrence, M.,
  O'Kelly, M., Tamayo, P., Weir, B.~A., Gabriel, S., Winckler, W., Gupta, S.,
  Jakkula, L., Feiler, H.~S., Hodgson, J.~G., James, C.~D., Sarkaria, J.~N.,
  Brennan, C., Kahn, A., Spellman, P.~T., Wilson, R.~K., Speed, T.~P., Gray,
  J.~W., Meyerson, M., Getz, G., Perou, C.~M., Hayes, D.~N., and {Cancer Genome
  Atlas Research Network} (2010), \enquote{Integrated {G}enomic {A}nalysis
  {I}dentifies {C}linically {R}elevant {S}ubtypes of {G}lioblastoma
  {C}haracterized by {A}bnormalities in {P}{D}{G}{F}{R}{A}, {I}{D}{H}1,
  {E}{G}{F}{R}, and {N}{F}1.} \textit{Cancer Cell}, 17, 98--110.

\bibitem[{Wang et~al.(2008)Wang, Shen, and Liu}]{WangShenLiu2008}
Wang, J., Shen, X., and Liu, Y. (2008), \enquote{Probability {E}stimation for
  {L}arge {M}argin {C}lassifiers,} \textit{Biometrika}, 95, 149--167.

\bibitem[{Wang and Shen(2007)}]{Wang2007}
Wang, L. and Shen, X. (2007), \enquote{{On $L_1$-norm Multi-class Support
  Vector Machines: Methodology and Theory},} \textit{Journal of the American
  Statistical Association}, 102, 595--602.

\bibitem[{Wegkamp(2007)}]{wegkamp2007lasso}
Wegkamp, M.~H. (2007), \enquote{{Lasso Type Classifiers with a Reject Option},}
  \textit{Electronic Journal of Statistics}, 1, 155--168.

\bibitem[{Wegkamp and Yuan(2011)}]{wegkamp2011support}
Wegkamp, M.~H. and Yuan, M. (2011), \enquote{{Support Vector Machines with a
  Reject Option},} \textit{Bernoulli}, 17, 1368--1385.

\bibitem[{Wu et~al.(2010)Wu, Zhang, and Liu}]{multiprob}
Wu, Y., Zhang, H.~H., and Liu, Y. (2010), \enquote{Robust {M}odel-free
  {M}ulticlass {P}robability {E}stimation,} \textit{Journal of the American
  Statistical Association}, 105, 424--436.

\bibitem[{Yuan and Wegkamp(2010)}]{yuan2010classification}
Yuan, M. and Wegkamp, M.~H. (2010), \enquote{{Classification Methods with
  Reject Option Based on Convex Risk Minimization},} \textit{Journal of Machine
  Learning Research}, 11, 111--130.

\bibitem[{Zhang and Liu(2013)}]{MLUM}
Zhang, C. and Liu, Y. (2013), \enquote{Multicategory {L}arge-margin {U}nified
  {M}achines,} \textit{Journal of Machine Learning Research}, 14, 1349--1386.

\bibitem[{Zhang and Liu(2014)}]{MAC}
--- (2014), \enquote{{Multicategory Angle-based Large-margin Classification},}
  \textit{Biometrika}, 101, 625--640.

\bibitem[{Zhang et~al.(2013)Zhang, Liu, and Wu}]{refit}
Zhang, C., Liu, Y., and Wu, Z. (2013), \enquote{{On the Effect and Remedies of
  Shrinkage on Classification Probability Estimation},} \textit{The American
  Statistician}, 67, 134--142.

\bibitem[{Zhang(2004)}]{zhangtong04}
Zhang, T. (2004), \enquote{Statistical {B}ehavior and {C}onsistency of
  {C}lassification {M}ethods {B}ased on {C}onvex {R}isk {M}inimization,}
  \textit{Annals of Statistics}, 32, 56--85.

\bibitem[{Zhao et~al.(2012)Zhao, Zeng, Rush, and Kosorok}]{zhao2012estimating}
Zhao, Y., Zeng, D., Rush, A.~J., and Kosorok, M.~R. (2012),
  \enquote{{Estimating Individualized Treatment Rules using Outcome Weighted
  Learning},} \textit{Journal of the American Statistical Association}, 107,
  1106--1118.

\bibitem[{Zhou(2002)}]{zhou2002covering}
Zhou, D.-X. (2002), \enquote{{The Covering Number in Learning Theory},}
  \textit{Journal of Complexity}, 18, 739--767.

\bibitem[{Zhu and Hastie(2005)}]{Zhu05}
Zhu, J. and Hastie, T.~J. (2005), \enquote{Kernel {L}ogistic {R}egression and
  the {I}mport {V}ector {M}achine,} \textit{Journal of Computational and
  Graphical Statistics}, 14, 185--205.

\end{thebibliography}



\section*{Supplementary material}

# Supplementary Materials of "On Reject and Refine Options in Multicategory Classification"

Chong Zhang, Wenbo Wang and Xingye Qiao

January 9, 2017

## 1 Proofs to Propositions and Theorems

Before the proofs, we first introduce a lemma for simplicity and completeness of further arguments.

**Lemma 1** (Zhang and Liu, 2014, Lemma 1). *Suppose we have an arbitrary $\boldsymbol{f} \in \mathbb{R}^{k-1}$. For any $u, v \in \{1, \ldots, k\}$ such that $u \neq v$, define $\boldsymbol{T}_{u,v} = \mathcal{Y}_u - \mathcal{Y}_v$. For any scalar $z \in \mathbb{R}$, $\langle (\boldsymbol{f} + z\boldsymbol{T}_{u,v}), \mathcal{Y}_w \rangle = \langle \boldsymbol{f}, \mathcal{Y}_w \rangle$, where $w \in \{1, \ldots, k\}$ and $w \neq u, v$. Furthermore, we have that $\langle (\boldsymbol{f} + z\boldsymbol{T}_{u,v}), \mathcal{Y}_u \rangle - \langle \boldsymbol{f}, \mathcal{Y}_u \rangle = -\langle (\boldsymbol{f} + z\boldsymbol{T}_{v,u}), \mathcal{Y}_v \rangle + \langle \boldsymbol{f}, \mathcal{Y}_v \rangle$.*

Lemma 1 shows that we can increase $\langle \boldsymbol{f}^*, \mathcal{Y}_i \rangle$ by an arbitrary $\epsilon > 0$, and decrease $\langle \boldsymbol{f}^*, \mathcal{Y}_j \rangle$ by the same $\epsilon$ without changing $\langle \boldsymbol{f}^*, \mathcal{Y}_l \rangle$ for $l \notin \{i, j\}$.

**Proof of Proposition 1**: We aim to find $\boldsymbol{f}^*$ that minimizes the conditional expected loss

$$\sum_{j=1}^{k} P_j \{ \sum_{i \neq y} \ell(\langle \boldsymbol{f}, \mathcal{Y}_i \rangle) \},$$

which is equivalent to find

$$\operatorname*{argmin}_{\boldsymbol{f}} \sum_{j=1}^{k} \ell(\langle \boldsymbol{f}, \mathcal{Y}_j \rangle)(1 - P_j).$$

We assume $P_1 \geq P_2 \geq \cdots \geq P_k$ in this proof for simplicity.

First, we show that $\langle \boldsymbol{f}^*, \mathcal{Y}_1 \rangle \geq \langle \boldsymbol{f}^*, \mathcal{Y}_2 \rangle \geq \cdots \langle \boldsymbol{f}^*, \mathcal{Y}_k \rangle$. We prove this by contradiction. Suppose $\langle \boldsymbol{f}^*, \mathcal{Y}_1 \rangle < \langle \boldsymbol{f}^*, \mathcal{Y}_2 \rangle$. By Lemma 1, we can define $\tilde{\boldsymbol{f}}$ such that $\langle \boldsymbol{f}^*, \mathcal{Y}_1 \rangle = \langle \tilde{\boldsymbol{f}}, \mathcal{Y}_2 \rangle$ and $\langle \boldsymbol{f}^*, \mathcal{Y}_2 \rangle = \langle \tilde{\boldsymbol{f}}, \mathcal{Y}_1 \rangle$. One can verify that $\sum_{j=1}^{k} \ell(\langle \boldsymbol{f}^*, \mathcal{Y}_j \rangle)(1 - P_j) > \sum_{j=1}^{k} \ell(\langle \tilde{\boldsymbol{f}}, \mathcal{Y}_j \rangle)(1 - P_j)$, which is a contradiction to the definition of $\boldsymbol{f}^*$. Notice that this

1

argument holds true for any pairwise comparisons between $\langle \boldsymbol{f}^*, \mathcal{Y}_i \rangle$ and $\langle \boldsymbol{f}^*, \mathcal{Y}_j \rangle$ for $i \neq j$. Therefore, we have $\langle \boldsymbol{f}^*, \mathcal{Y}_1 \rangle \geq \langle \boldsymbol{f}^*, \mathcal{Y}_2 \rangle \geq \cdots \langle \boldsymbol{f}^*, \mathcal{Y}_k \rangle$.

The second step is to start from $\boldsymbol{f} = 0$, and consider the pairwise comparison between $(P_1, \langle \boldsymbol{f}, \mathcal{Y}_1 \rangle)$ and $(P_q, \langle \boldsymbol{f}, \mathcal{Y}_j \rangle)$ for $q > 1$, in order to decrease the conditional expected loss. By Lemma 1 and similar argument as in Section 3.1, one can verify that if $a(1 - P_1) < (1 - P_q)$, we should increase $\langle \boldsymbol{f}, \mathcal{Y}_1 \rangle$ and decrease $\langle \boldsymbol{f}, \mathcal{Y}_q \rangle$ to decrease the conditional expected loss. If $a(1 - P_1) \geq (1 - P_q)$, we should keep $\langle \boldsymbol{f}, \mathcal{Y}_q \rangle$ at 0. After $k - 1$ such comparisons, one can verify that $\boldsymbol{f}$ is such that if the assumption in Proposition 1 holds, then $\langle \boldsymbol{f}, \mathcal{Y}_1 \rangle > 0$, $\langle \boldsymbol{f}, \mathcal{Y}_2 \rangle = \cdots = \langle \boldsymbol{f}, \mathcal{Y}_j \rangle = 0$, and $\langle \boldsymbol{f}, \mathcal{Y}_q \rangle < 0$ for $j + 1 \leq q \leq k$. Note that the fact $\ell'(u)$ is a constant for $u > 0$ is essential for this sequence of pairwise comparisons to hold.

The last step is to check that $\boldsymbol{f}^* = \boldsymbol{f}$, where $\boldsymbol{f}$ is obtained in the second step. To this end, notice that if $\boldsymbol{f}^* \neq \boldsymbol{f}$, we can always perform the pairwise comparison as in the second step to decrease the conditional expected loss. Therefore, Proposition 1 holds. □

**Proof of Proposition 2**: For any fixed $\boldsymbol{x}$, the conditional expected loss for the reject option is $d$. To predict class label, clearly $\hat{y} = Y_{(1)}$ is the only admissible decision, whose conditional expected loss is $1 - P_{(1)}$. Therefore, we would predict the label when $1 - P_{(1)} < d$, and we reject when $1 - P_{(1)} \geq d$. □

**Proof of Proposition 3**: We assume $P_1 \geq P_2 \geq \cdots \geq P_k$ in this proof for simplicity. To prove the lower bound, suppose $a(1 - P_1) > (1 - P_k)$. Consequently, we have $a(1 - P_1) > (1 - P_j)$ for $j \geq 2$, which further leads to $a(k-1)(1 - P_1) > k - 1 - (1 - P_1)$. With some calculation, this is equivalent to $1 - P_1 > \frac{k-1}{a(k-1)+1}$. Therefore, by letting $\frac{k-1}{a(k-1)+1} = d$, or equivalently, $a = a_1$, we can prove that the lower bound inequality holds.

To prove the upper bound, suppose $1 - P_1 > d$. With $a = a_2 = \frac{(k-1)(1-d)}{d}$, we have $a(1 - P_1) > (k-1)(1-d) > (k-1)P_1 > \sum_{j=1}^{k-1} P_j = 1 - P_k$. This proves the upper bound.

To see the tightness of these 2 bounds, one can easily construct numerical counter examples, and we omit the details here. □

**Proof of Theorem 1**: The key to the proof of this theorem is to bound the tail probability that the deviation of a related empirical process from its expected value exceeds a certain threshold. This consists of two major parts. The first part is to transform the problem into the empirical process, and the second part is to bound the corresponding tail probability.

Recall the definition of $\mathcal{F}(p, k, s)$. Define $t(p, s) = s$ if we use the $L_1$ penalty, and

$t(p, s) = (ps)^{1/2}$ if we use the $L_2$ penalty. One can verify that for $L_1$ or $L_2$ penalized method, and any $j \in \{1, \ldots, k-1\}$, $|\hat{f}_j| = |\hat{\boldsymbol{\beta}}_j^T \boldsymbol{x}| \leq t(p, s)$. Therefore, in future arguments, it suffices to consider $\tilde{\mathcal{F}}(p, k, s) = \mathcal{F}(p, k, s) \cap \{\boldsymbol{f} : \|\boldsymbol{f}\| \leq (k-1)t(p, s)\}$. Furthermore, define $\boldsymbol{f}^{(p,k,s)} = \mathrm{argmin}_{\boldsymbol{f} \in \tilde{\mathcal{F}}(p,k,s)} \mathbb{E}[\sum_{j \neq y} \ell\{\langle \boldsymbol{f}, \mathcal{Y}_j \rangle\}]$,

$$h_{\boldsymbol{f}}(\cdot) = \{2(k-1)\frac{1-d}{d}t(p,s)\}^{-1}\{\sum_{j \neq \cdot} \ell(\langle \boldsymbol{f}, \mathcal{Y}_j \rangle) - \sum_{j \neq \cdot} \ell(\langle \boldsymbol{f}^{(p,k,s)}, \mathcal{Y}_j \rangle)\},$$

and $\bar{H} = \{h_{\boldsymbol{f}} : \boldsymbol{f} \in \tilde{\mathcal{F}}(p, k, s)\}$. Since $\ell$ is Lipschitz with Lipschitz constant $\frac{(k-1)(1-d)}{d}$ and $\sum_{j=1}^k \langle \boldsymbol{f}, \mathcal{Y}_j \rangle = 0$, $|\sum_{j \neq \cdot} \ell(\langle \boldsymbol{f}, \mathcal{Y}_j \rangle) - \sum_{j \neq \cdot} \ell(\langle \boldsymbol{f}', \mathcal{Y}_j \rangle)| \leq |\frac{(k-1)(1-d)}{d}\langle \boldsymbol{f} - \boldsymbol{f}', \cdot \rangle| \leq 2\frac{(k-1)(1-d)}{d}t(p, s)$. Therefore, we have the $L_2(Q)$ diameter of $\{\sum_{j \neq \cdot} \ell(\langle \boldsymbol{f}, \mathcal{Y}_j \rangle) - \sum_{j \neq \cdot} \ell(\langle \boldsymbol{f}^{(p,k,s)}, \mathcal{Y}_j \rangle)\}$ is bounded by $\{2\frac{(k-1)(1-d)}{d}t(p, s)\}$, and the $L_2(Q)$ diameter of $\bar{H}$ is bounded by 1. Here $Q$ is any arbitrary distribution.

The next lemma bounds the complexity of $\bar{H}$ in terms of its $L_2(Q)$ entropy. For any $\epsilon > 0$, we can define $\mathcal{G}$ to be an $\epsilon$-net of a function class $\mathcal{F}$ if, for any $f \in \mathcal{F}$, there exists $g \in \mathcal{G}$ such that $\|g - f\|_{Q,2} \leq \epsilon$. Let the $L_2(Q)$ covering number $N\{\epsilon, \mathcal{F}, L_2(Q)\}$ be the minimum size of all such possible $\epsilon$-nets, and denote by $H\{\epsilon, \mathcal{F}, L_2(Q)\}$ the logarithm of $N\{\epsilon, \mathcal{F}, L_2(Q)\}$, which is referred to as the $L_2(Q)$ entropy. Define the uniform $L_2(Q)$ covering number, $N(\epsilon, \mathcal{F})$, to be $\sup_Q N\{\epsilon, \mathcal{F}, L_2(Q)\}$, and define the uniform $L_2(Q)$ entropy $H(\epsilon, \mathcal{F})$ in a similar manner. Lemma 2 gives an upper bound on $H(\epsilon, \bar{H})$.

**Lemma 2.** *For any $\epsilon > 0$, $H(\epsilon, \bar{H}) \leq \frac{2(k-1)}{\epsilon^2} \log(e + 2pe\epsilon^2)$.*

**Proof of Lemma 2**: To bound the $L_2(Q)$ entropy of $\bar{H}$, we can first bound the $L_2(Q)$ entropy of $\mathcal{G} := \{\sum_{j \neq \cdot} \ell(\langle \boldsymbol{f}, \mathcal{Y}_j \rangle) : \sum_{j=1}^{k-1} \|\boldsymbol{\beta}_j\|_1 \leq t(p, s)\}$, as a $\{2(k-1)\frac{1-d}{d}t(p,s)\epsilon\}$-net on $\mathcal{G}$ naturally introduces an $\epsilon$-net on $\bar{H}$. To this end, we find an $\epsilon$-net on $\mathcal{G}$. Let $g = \sum_{j \neq \cdot} \ell(\langle \boldsymbol{f}, \mathcal{Y}_j \rangle)$, $g' = \sum_{j \neq \cdot} \ell(\langle \boldsymbol{f}', \mathcal{Y}_j \rangle) \in \mathcal{G}$. Notice that

$$\|g - g'\|_{Q,2}^2 = \mathbb{E}[\sum_{j \neq \cdot} \ell\{\langle \mathcal{Y}_j, \boldsymbol{f}(\boldsymbol{X}) \rangle\} - \sum_{j \neq \cdot} \ell\{\langle \mathcal{Y}_j, \boldsymbol{f}'(\boldsymbol{X}) \rangle\}]^2$$

$$\leq \mathbb{E}\{\frac{(k-1)(1-d)}{d} \sum_{j \neq \cdot} \langle \mathcal{Y}_j, \boldsymbol{f}(\boldsymbol{X}) - \boldsymbol{f}'(\boldsymbol{X}) \rangle\}^2$$

$$\leq \mathbb{E}\{\frac{(k-1)(1-d)}{d} \sum_{j=1}^{k-1} |f_j(\boldsymbol{X}) - f'_j(\boldsymbol{X})|\}^2$$

$$\leq \frac{(k-1)^3(1-d)^2}{d^2} \sum_{j=1}^{k-1} \|f_j - f'_j\|_{Q,2}^2.$$

where the last step is from the Cauchy-Schwartz inequality. Next, we define $\vec{\boldsymbol{x}} = (\boldsymbol{x}_1^T, \ldots, \boldsymbol{x}_{k-1}^T)^T$

3

with each $\boldsymbol{x}_j$ a $p$-dimensional vector. Let $\vec{f}(\vec{\boldsymbol{x}}) = \sum_{j=1}^{k-1} \boldsymbol{\beta}_j^T \boldsymbol{x}_j$. Also let $\vec{Q}$ be the distribution of $\vec{\boldsymbol{X}} = (\delta_1 \boldsymbol{X}_1, \ldots, \delta_{k-1} \boldsymbol{X}_{k-1})$, where $\boldsymbol{X}_j$'s are independent and identically distributed with any arbitrary distribution $Q$, and $(\delta_1, \ldots, \delta_{k-1})$ has a joint distribution $\text{pr}\{(\delta_1, \ldots, \delta_{k-1})^T = e_j\} = (k-1)^{-1}$. Thus we may conclude that $\sum_{j=1}^{k-1} \|f_j - f_j'\|_{Q,2}^2 = (k-1)\mathbb{E}_{\vec{Q}}(\vec{f} - \vec{f}')^2$, and $\|g - g'\|_{Q,2}^2 \leq \frac{(k-1)^4(1-d)^2}{d^2} \|\vec{f} - \vec{f}'\|_{Q,2}^2$. Consequently, if we can bound $L_2(Q)$ entropy of the function class $\vec{\mathcal{F}} = \{\vec{f} : \vec{f}(\vec{\boldsymbol{x}}) = \sum_{j=1}^{k-1} \sum_{l=1}^{p} \beta_{j,l} x_{j,l}; \sum_{j=1}^{k-1} \|\boldsymbol{\beta}_j\|_1 \leq t(p,s)\}$, we can bound $\bar{H}$.

To bound the entropy of $\vec{\mathcal{F}}$, we define $w_{j,l}(\vec{\boldsymbol{x}}) = t(p,s)x_{j,l}$. Hence, $\mathcal{J} = \{\pm w_{j,l}\}$ forms a basis for $\vec{\mathcal{F}}$. In other words, each $\vec{f} = \sum_{j=1}^{k-1} \sum_{l=1}^{p} \beta_{j,l} x_{j,l} = \sum_{j=1}^{k-1} \sum_{l=1}^{p} |\beta_{j,l}|\{\text{sign}(\beta_{j,l}) w_{j,l}(\vec{\boldsymbol{x}})\}/t(p,s)$ is a convex combination of $w_{j,l}$. Thus, $\vec{\mathcal{F}}$ is the convex hull of $\mathcal{J}$. By Lemma 2.6.11 in van der Vaart and Wellner (2000), $N\{\epsilon \text{diam}\mathcal{J}, \vec{\mathcal{F}}, L_2(\vec{Q})\} \leq [e + e\{2p(k-1)\}\epsilon^2]^{2/\epsilon^2}$, where $\text{diam}\mathcal{J} = \sup_{J_1, J_2 \in \mathcal{J}} \|J_1 - J_2\|_{\vec{Q},2} \leq 2t(p,s)$. Thus, we conclude that $N\{\epsilon, \bar{H}, L_2(Q)\} = N\{2(k-1)\frac{1-d}{d}t(p,s)\epsilon, \mathcal{G}, L_2(Q)\} \leq N\{2t(p,s)\sqrt{(k-1)^{-1}}\epsilon, \mathcal{J}, L_2(\vec{Q})\} \leq (e + 2pe\epsilon^2)^{2(k-1)/\epsilon^2}$. Since the final bound is independent of $Q$, we have that the bound is uniform for any $Q$. ∎

The next lemma shows that in order to show the result in Theorem 1, we can focus on bounding a tail probability.

**Lemma 3.** *For given $n$, $p$, $k$, and $s$, assume that there exists $M > 0$ that satisfies*

$$(\log_2 \frac{16\sqrt{6}\epsilon_0}{M} + 1)^2 \left\{ \frac{256 \log(e + 2pe\epsilon_0^2)}{n} \right\} \leq \frac{M^2}{256}, \tag{1}$$

*where $\epsilon_0 > 0$ is such that*

$$\frac{2(k-1)\log(e + 2pe\epsilon_0^2)}{\epsilon_0^2} = \frac{1}{4} nM^2. \tag{2}$$

*Then for $d_{n,p,k} = \inf_{\boldsymbol{f} \in \mathcal{F}(p,k,s)} e_\ell(\boldsymbol{f}, \boldsymbol{f}^{(p,k)})$, we have*

$$\text{pr}\{e_\ell(\hat{\boldsymbol{f}}, \boldsymbol{f}^{(p,k)}) \geq 8(k-1)t(p,s)M + d_{n,p,k}\} \leq 6(1 - \frac{1}{16nM^2})^{-1} \exp(-nM^2).$$

**Proof of Lemma 3**: Define the empirical process $h \to P_n h - Ph$, where $h \in \bar{H}$, $Ph = \int h d\mathbb{P}$ and $P_n h = n^{-1} \sum_{i=1}^{n} h(y_i)$. We have, by definition of $d_{n,p,k}$,

$$\text{pr}\{e_\ell(\hat{\boldsymbol{f}}, \boldsymbol{f}^{(p,k)}) > 8(k-1)t(p,s)M + d_{n,p,k}\} \leq \text{pr}[e_\ell(\hat{\boldsymbol{f}}, \boldsymbol{f}^{(p,k,s)})\{2(k-1)t(p,s)\}^{-1} > 4M].$$

Since $\hat{\boldsymbol{f}}$ is such that $e_\ell(\hat{\boldsymbol{f}}, \boldsymbol{f}^{(p,k,s)})\{2(k-1)t(p,s)\}^{-1} > 4M$, and $\hat{\boldsymbol{f}}$ minimizes the em-

4

pirical loss $n^{-1}\sum_{i=1}^{n}\sum_{j\neq y_i}\ell\{\langle \boldsymbol{f}(\boldsymbol{x}_i),\mathcal{Y}_j\rangle\}$, we have $n^{-1}\sum_{i=1}^{n}\{\sum_{j\neq y_i}\ell(\langle \boldsymbol{f}^{(p,k,s)},\mathcal{Y}_j\rangle) - \sum_{j\neq y_i}\ell(\langle \hat{\boldsymbol{f}},\mathcal{Y}_j\rangle)\} \geq 0$. Hence,

$$\mathrm{pr}\{e_\ell(\hat{\boldsymbol{f}},\boldsymbol{f}^{(p,k)}) > 8(k-1)t(p,s)M + d_{n,p,k}\}$$

$$\leq \mathrm{pr}^{outer}\left[\sup_{\boldsymbol{f}\in\tilde{\mathcal{F}}(p,k,s):e_\ell(\boldsymbol{f},\boldsymbol{f}^{(p,k,s)})\{2(k-1)t(p,s)\}^{-1}>4M}\frac{1}{n}\sum_{i=1}^{n}[\sum_{j\neq y_i}\{\ell(\langle \boldsymbol{f}^{(p,k,s)},\mathcal{Y}_j\rangle) - \ell(\langle \boldsymbol{f},\mathcal{Y}_j\rangle)\}] > 0\right]$$

$$\leq pr^{outer}\Big[\sup_{\boldsymbol{f}\in\tilde{\mathcal{F}}(p,k,s):e_\ell(\boldsymbol{f},\boldsymbol{f}^{(p,k,s)})\{2(k-1)t(p,s)\}^{-1}>4M}-\frac{1}{n}\sum_{i=1}^{n}[h_{\boldsymbol{f}}(y_i) - \mathbb{E}\{h_{\boldsymbol{f}}(Y)\}]$$

$$> \{2(k-1)t(p,s)\}^{-1}\mathbb{E}\{\sum_{j\neq y}\ell(\langle \boldsymbol{f},\mathcal{Y}_j\rangle) - \sum_{j\neq y}\ell(\langle \boldsymbol{f}^{(p,k,s)},\mathcal{Y}_j\rangle)\}\Big].$$

Here $\mathrm{pr}^{outer}$ is the outer probability. In the region $\boldsymbol{f}\in\tilde{\mathcal{F}}(p,k,s) : e_\ell(\boldsymbol{f},\boldsymbol{f}^{(p,k,s)})\{2(k-1)t(p,s)\}^{-1} > 4M$, $\{2(k-1)t(p,s)\}^{-1}\mathbb{E}\{\sum_{j\neq y}\ell(\langle \boldsymbol{f},\mathcal{Y}_j\rangle) - \sum_{j\neq y}\ell(\boldsymbol{f}^{(p,k,s)},\mathcal{Y}_j)\}$ is always larger than $4M$. Hence we have

$$\mathrm{pr}\{e_\ell(\hat{\boldsymbol{f}},\boldsymbol{f}^{(p,k)}) > 8(k-1)t(p,s)M + d_{n,p,k}\} \leq \mathrm{pr}^{outer}(\sup_{h\in\bar{H}}|P_nh - Ph| > 4M).$$

The rest part of the proof is to bound the tail probability $\sup_{h\in\bar{H}}|P_nh - Ph| > 4M$. Notice that the entropy of $\bar{H}$ is given in Lemma 2, and the entropy is of the order $\epsilon^{-2}$. Thus, by (1), (2), and Theorem A.2 in Wang and Shen (2007), we have $\mathrm{pr}^{outer}(\sup_{h\in\bar{H}}|P_nh - Ph| > 4M) \leq 6\{1-(1/16nM^2)\}^{-1}\exp(-nM^2)$, and this completes the proof. ∎

With Lemma 3 proved, we can proceed to prove Theorem 1.

Let $M = 5r\log(r^{-1})$. We need to verify that (1) holds for the choice of $M$ and $\epsilon_0$ in (2). First, note that $\epsilon_0$ goes to 0. Because if $\epsilon_0$ is bounded away from 0 and $\infty$, the left hand side of (2) is of order $O(\log p)$, and the right hand side of (2) is of order $O\{\log p\log^2(r)^{-1}\}$, which is a contradiction. If $\epsilon_0 \to \infty$, the left hand side of (2) is of order $o(\log p)$, which is still a contradiction. Next, note that (1) is equivalent to

$$(\log_2\frac{16\sqrt{6}\epsilon_0}{M} + 1)^2 \leq \frac{nM^2}{2^{16}\log(e+2pe\epsilon_0^2)}. \tag{3}$$

We have $\log_2(16\sqrt{6}\epsilon_0)/M + 1 \propto \log_2(\epsilon_0/M) \preceq \log_2(1/M) \preceq \log(1/r)$, where $\propto$ means "equivalent up to a constant", and $\preceq$ means "less than or equal to up to a constant". As a result, the left hand side of (3) has an order no greater than $O\{\log^2(r^{-1})\}$. For the right hand side of (3), we have $\epsilon_0^{-2} \preceq (nM^2)/\{2^{16}\log(e+2pe\epsilon_0^2)\}$. The left hand side of (2) has order $O\{\log p\log^2(r)^{-1}\}$. If the order of $1/\epsilon_0$ is less than that of $\log(1/r)$, we have the

order of the right hand side of (2) smaller than $O\{\log p \log^2(1/r)\}$, because $\epsilon_0$ goes to 0. Thus, (1) is valid, because the order of left hand side of (3) is less than that of the right hand side.

Finally, note that $nM^2 = 25 \log p \log^2(1/r) > 2.5 \log p \log(1/r) \geq 2.5 \log n > 2 \log n$. Hence, we have $\exp(-nM^2) \leq \exp(-2 \log n) = n^{-2}$. The desired result in Theorem 1 then follows from the Borel-Cantelli Lemma. □

**Proof of Theorem 2**: The key to the proof is to show that with any kernel function such that $K(\cdot, \cdot) \leq \infty$, the corresponding entropy number of the function space is approximately in the order $\epsilon^{-2}$.

Let $t(p, s) = s$ for kernel learning. Define $\boldsymbol{f}^{(p,k,s)}$, $h_{\boldsymbol{f}}(\cdot)$, and $\bar{H}$ in a similar manner with respect to the linear learning case in the proof of Theorem 1. Here without loss of generality, assume that the kernel function is upper bounded by 1. Note that the theory can be naturally generalized to other cases with different upper bounds. Now, with the assumption that the kernel is separable, one can verify that the $L_2$ diameter of $\bar{H}$ can be bounded by 1. Next, instead of bounding the uniform entropy as in the linear case, we bound the empirical uniform entropy for kernel learning. In particular, let $T_X$ be the empirical measure of a training data set $\{(\boldsymbol{x}_1, y_1), \ldots, (\boldsymbol{x}_m, y_m)\}$, and let the $L_2$ norm be defined as $\|f\|_{L_2(T_X)} = \left(\frac{1}{m} \sum_{i=1}^{m} |f(\boldsymbol{x}_i, y_i)|^2\right)^{1/2}$. We can define the $L_2(T_X)$ covering number and entropy number in an obvious manner. In kernel learning, let $H(\epsilon, \bar{H})$ be $\sup_{T_X} H(\epsilon, \bar{H}, L_2(T_X))$, which we call the empirical uniform entropy. Next, we bound $H(\epsilon, \bar{H})$. Notice that $C$ is a constant that may change in different context.

**Lemma 4.** *For any $\epsilon > 0$, $H(\epsilon, \bar{H}) \leq C \epsilon^{-2} \log(\frac{1}{\epsilon})$.*

**Proof of Lemma 4**: Let $\mathcal{G} := \{\sum_{j \neq \cdot} \ell(\langle \boldsymbol{f}, \mathcal{Y}_j \rangle) : \sum_{j=1}^{k-1} J(\boldsymbol{f}) \leq s\}$. Let $g$ and $g'$ be defined as in the proof of Lemma 2. One can verify that

$$\|g - g'\|_{L_2(T_X)}^2 = \mathbb{E}[\sum_{j \neq \cdot} \ell\{\langle \mathcal{Y}_j, \boldsymbol{f}(\boldsymbol{X})\rangle\} - \sum_{j \neq \cdot} \ell\{\langle \mathcal{Y}_j, \boldsymbol{f}'(\boldsymbol{X})\rangle\}]^2$$
$$\leq \mathbb{E}\{\frac{(k-1)(1-d)}{d} \sum_{j \neq \cdot} \langle \mathcal{Y}_j, \boldsymbol{f}(\boldsymbol{X}) - \boldsymbol{f}'(\boldsymbol{X})\rangle\}^2$$
$$\leq \frac{(k-1)^2(1-d)^2}{d^2} \mathbb{E}\{\sum_{j=1}^{k-1} |f_j(\boldsymbol{X}) - f_j'(\boldsymbol{X})|\}^2.$$

Hence, the $L_2(T_X)$ covering number of $\mathcal{G}$ can be upper bounded through bounding the $L_2(T_X')$ covering number of $\mathcal{G}'$, which is a set that ranges over all individual classification functions whose norm is upper bounded by $s$. Here $T_X'$ is the empirical measure of

6

$\{\delta_1 \boldsymbol{X}, \delta_2 \boldsymbol{X}, \ldots, \delta_{k-1} \boldsymbol{X}\}$, where $\boldsymbol{X} = (\boldsymbol{x}_1, \ldots, \boldsymbol{x}_n)$, and $(\delta_1, \ldots, \delta_{k-1})$ has a joint distribution $\text{pr}\{(\delta_1, \ldots, \delta_{k-1})^T = e_j\} = (k-1)^{-1}$. Next, by similar arguments as in the proof of Lemma 2 in Zhang et al. (2016), we have $\sup_{T_X} N(\epsilon, \mathcal{G}, L_2(T_X)) \leq \frac{5 \exp(C\epsilon^{-2})}{\epsilon}$. Therefore, the claim in Lemma 4 holds. ∎

The rest of the proof is to notice that the order of the entropy number is $\epsilon^{-2} \log(1/\epsilon)$, which is very close to $\epsilon^{-2}$. Hence, one can verify that (1) and (2) hold in general. By similar arguments as in the proof of linear learning, we can prove Theorem 2. □

**Proof of Theorem 3**: First, notice that for $j \neq 1$, $(1 - P_{(j)})\ell'(\langle \mathcal{Y}_j, \boldsymbol{f}^* \rangle) = a(1 - P_{(1)})$. Hence, as we assume that the probabilities are bounded away from 0, we can conclude that for a fixed loss function, $\langle \mathcal{Y}_{(j)}, \boldsymbol{f}^* \rangle$ is bounded away from $\infty$ for all $j \neq 1$. Consequently, we have that $\ell''(\langle \mathcal{Y}_{(j)}, \boldsymbol{f}^* \rangle)$ has a lower bound for any $\langle \mathcal{Y}_j, \boldsymbol{f}^* \rangle < 0$. Denote by $\zeta$ this lower bound. Next, define $r(\boldsymbol{f}) = \sum_{j=1}^k (1 - P_j)\{\ell(\langle \boldsymbol{f}, \mathcal{Y}_j \rangle) - \ell(\langle \boldsymbol{f}^*, \mathcal{Y}_j \rangle)\}$ for any $\boldsymbol{f}$. For brevity in expression, we let $\boldsymbol{f}^* \in \mathcal{F}(p, k)$. Notice that if $\boldsymbol{f}^* \notin \mathcal{F}(p, k)$, the proof becomes slightly more complicated in the approximation error term. Hence we have $r(\hat{\boldsymbol{f}}) \geq 0$, and $\nabla r(\boldsymbol{f}) |_{\boldsymbol{f}^*} = \boldsymbol{0}$.

Without loss of generality, assume that $\eta_0$ is small enough such that $a\eta_0(k-1) < 1$. Define $\rho(\boldsymbol{f}, \boldsymbol{f}^*) = \frac{\zeta}{2ak} \max\left(1, \frac{(k-1)\eta_0}{1/a - (k-1)\eta_0}\right) \sum_{j=1}^k (\langle \boldsymbol{f}, \mathcal{Y}_j \rangle - \langle \boldsymbol{f}^*, \mathcal{Y}_j \rangle)^2$. For $a(1 - P_{(1)}) > (1 - P_{(k)})$ and $\boldsymbol{f}$ close to $\boldsymbol{f}^*$, we have by Taylor's expansion, $r(\boldsymbol{f}) \geq \{(1 - P_{(1)}) - \frac{1}{a}(1 - P_{(k)})\} \sum_{j=1}^k \frac{\ell''(\langle \boldsymbol{f}^*, \mathcal{Y}_j \rangle)}{2} (\langle \boldsymbol{f}, \mathcal{Y}_j \rangle - \langle \boldsymbol{f}^*, \mathcal{Y}_j \rangle)^2$. Notice that $\sum_{j=1}^k \langle \boldsymbol{f}, \mathcal{Y}_j \rangle = \sum_{j=1}^k \langle \boldsymbol{f}^*, \mathcal{Y}_j \rangle = 0$, and we can conclude that $r(\boldsymbol{f}) \geq \frac{\zeta}{2k}\{(1 - P_{(1)}) - \frac{1}{a}(1 - P_{(k)})\} \sum_{j=1}^k (\langle \boldsymbol{f}, \mathcal{Y}_j \rangle - \langle \boldsymbol{f}^*, \mathcal{Y}_j \rangle)^2 \geq |a(1 - P_{(1)}) - (1 - P_{(k)})|\rho(\boldsymbol{f}, \boldsymbol{f}^*)$. On the other hand, if $a(1 - P_{(1)}) < (1 - P_{(k)})$, one can verify that $\frac{1}{a}(1 - P_{(k)}) - (1 - P_{(1)}) \leq \frac{1}{a} - (k-1)\eta_0$. Hence, by similar argument as above, we have $r(\boldsymbol{f}) \geq \frac{\zeta}{2k}\{\frac{1}{a}(1 - P_{(k)}) - (1 - P_{(1)})\} \frac{(k-1)\eta_0}{1/a - (k-1)\eta_0} \geq |a(1 - P_{(1)}) - (1 - P_{(k)})|\rho(\boldsymbol{f}, \boldsymbol{f}^*)$.

Next, define $g_{\boldsymbol{f}}(\boldsymbol{x}, y) = \sum_{j \neq y} \ell\{\langle \boldsymbol{f}, \mathcal{Y}_j \rangle\} - \sum_{j \neq y} \ell\{\langle \boldsymbol{f}^*, \mathcal{Y}_j \rangle\}$. We prove that $\mathbb{P}g^2 \leq B(\mathbb{P}g)^{\alpha/(1+\alpha)}$ for a constant $B$ that does not depend on $n$. To this end, notice that for any $\boldsymbol{f}$,

$$\begin{aligned}
\mathbb{E}\{g_{\boldsymbol{f}}(\boldsymbol{x}, y)\} &= \mathbb{E}\{r(\boldsymbol{f})\} \\
&\geq \mathbb{E}\rho(\boldsymbol{f}, \boldsymbol{f}^*)|a(1 - P_{(1)}) - (1 - P_{(k)})| \\
&\geq t\mathbb{E}\{\rho(\boldsymbol{f}, \boldsymbol{f}^*)\}I_{|a(1-P_{(1)})-(1-P_{(k)})|\geq t} \\
&= t[\mathbb{E}\{\rho(\boldsymbol{f}, \boldsymbol{f}^*)\} - \mathbb{E}\{\rho(\boldsymbol{f}, \boldsymbol{f}^*)I_{|a(1-P_{(1)})-(1-P_{(k)})|<t}\}] \\
&\geq t[\mathbb{E}\{\rho(\boldsymbol{f}, \boldsymbol{f}^*)\} - C_1(s)t^\alpha],
\end{aligned}$$

where $C_1(s)$ is a linear function of $s$, such that $C_1(s) \geq \rho(\boldsymbol{f}, \boldsymbol{f}^*)$ for all $\boldsymbol{f}$. Choose

7

$t = \left[\frac{\mathbb{E}\{\rho(\boldsymbol{f},\boldsymbol{f}^*)\}}{2C_1(s)}\right]^{1/\alpha}$, and we have $\mathbb{E}\{\rho(\boldsymbol{f},\boldsymbol{f}^*)\} \leq C_2(s)[\mathbb{E}\{g_{\boldsymbol{f}}(\boldsymbol{x},y)\}]^{\alpha/(1+\alpha)}$, where $C_2(s)$ is another linear function of $s$.

On the other hand, notice that

$$\mathbb{E}\{g_{\boldsymbol{f}}(\boldsymbol{x},y)\}^2 = \mathbb{E}[\mathbb{E}\{g_{\boldsymbol{f}}(\boldsymbol{x},y)\}^2|\boldsymbol{x}]$$
$$\leq C_3\mathbb{E}\{\rho(\boldsymbol{f},\boldsymbol{f}^*)\},$$

where $C_3$ is a universal constant. Hence, combining the above inequalities to obtain that

$$\mathbb{E}\{g_{\boldsymbol{f}}(\boldsymbol{x},y)\}^2 \leq C_4(s)\mathbb{E}\{g_{\boldsymbol{f}}(\boldsymbol{x},y)\}^{\alpha/(1+\alpha)},$$

where $C_4(s)$ is a linear function of $s$.

Next, let $\mathbb{P}_n g_{\boldsymbol{f}} = \frac{1}{n}\sum_{i=1}^n g_{\boldsymbol{f}}(\boldsymbol{x}_i, y_i)$, and $\mathbb{P}g_{\boldsymbol{f}} = \mathbb{E}_{\mathbb{P}(\boldsymbol{x},y)}g_{\boldsymbol{f}}$. We have

$$e_\ell(\hat{\boldsymbol{f}}, \boldsymbol{f}^*) = \mathbb{E}\{2\mathbb{P}_n g_{\hat{\boldsymbol{f}}} + (\mathbb{P} - 2\mathbb{P}_n)g_{\hat{\boldsymbol{f}}}\}$$
$$\leq 2\mathbb{E}\{\inf_{\boldsymbol{f}\in\mathcal{F}(p,k,s)} 2\mathbb{P}_n g_{\boldsymbol{f}} + \sup_{\boldsymbol{f}\in\mathcal{F}(p,k,s)}(\mathbb{P}-2\mathbb{P}_n)g_{\boldsymbol{f}}\}$$
$$\leq 2\inf_{\boldsymbol{f}\in\mathcal{F}(p,k,s)}\mathbb{E}(\mathbb{P}_n g_{\boldsymbol{f}}) + \mathbb{E}\{\sup_{\boldsymbol{f}\in\mathcal{F}(p,k,s)}(\mathbb{P}-2\mathbb{P}_n)g_{\boldsymbol{f}}\}$$
$$\leq 2d_{n,p,k} + 2(k-1)\left[\epsilon_n + \mathrm{pr}\{\sup_{\boldsymbol{f}\in\mathcal{F}_n(p,k,s)}(\mathbb{P}-2\mathbb{P}_n)g_{\boldsymbol{f}} \geq \epsilon_n\}\right],$$

where $\mathcal{F}_n(p,k,s)$ is the space of functions that corresponds to an $\epsilon_n$-net of $\bar{H}$. Furthermore, because the entropy number of $\mathcal{F}_n(p,k,s)$ is the same as that of $\bar{H}$, and is of order $o(\epsilon_n^{-\delta})$ for any $\delta > 0$ (Zhou, 2002), we have, by Bernstein's Inequality,

$$\mathrm{pr}\{\sup_{\boldsymbol{f}\in\mathcal{F}_n(p,k,s)}(\mathbb{P}-2\mathbb{P}_n)g_{\boldsymbol{f}} \geq \epsilon_n\} \leq \sum_{\boldsymbol{f}\in\mathcal{F}_n(p,k,s)}\mathrm{pr}\{(\mathbb{P}-\mathbb{P}_n)g_{\boldsymbol{f}} \geq \frac{1}{2}(\mathbb{P}g_{\boldsymbol{f}}+\epsilon_n)\}$$
$$\leq |\mathcal{F}_n(p,k,s)|\max_{\boldsymbol{f}\in\mathcal{F}_n(p,k,s)}\exp\left\{-\frac{n}{8}\frac{(\mathbb{P}g_{\boldsymbol{f}}+\epsilon_n)^2}{\mathbb{P}g_{\boldsymbol{f}}^2 + C_1(s)(\mathbb{P}g_{\boldsymbol{f}}+\epsilon_n)/6}\right\}$$
$$\leq \exp(C_5\epsilon_n^{-\delta} - C_6(s)n\epsilon_n^{2-\alpha/(1+\alpha)}),$$

where $C_5$ is a universal constant, and $C_6(s)$ is a linear function of $s$.

Let $\epsilon_n = M(s)n^{-(1+\alpha)/(2+\alpha)}$, where $M(s)$ is a linear function of $s$. We choose $M(s)$ such that $C_5\epsilon_n^{-\delta} = \frac{1}{2}C_6(s)n\epsilon_n^{2-\alpha/(1+\alpha)}$ and $\exp(-n\epsilon_n^{2-\alpha/(1+\alpha)}) = o(\epsilon_n)$. We then have $e_\ell(\hat{\boldsymbol{f}}, \boldsymbol{f}^*) \leq 2d_{n,p,k} + 2C_7(k-1)sn^{-(1+\alpha)/(2+\alpha)}$ for a universal constant $C_7$. This completes the proof.

For binary loss functions that are flat for large enough $u$, one can verify that the

8

largest $|\langle \boldsymbol{f}^*, \mathcal{Y}_j \rangle|$ is bounded. In that case, it is possible to obtain similar results by defining $\rho(\boldsymbol{f}, \boldsymbol{f}^*)$ to be $C \sum_{j=1}^{k} |\langle \boldsymbol{f}, \mathcal{Y}_j \rangle - \langle \boldsymbol{f}^*, \mathcal{Y}_j \rangle|$, where $C$ can be chosen by careful analysis of the loss function. See Bartlett and Wegkamp (2008) for an example in the binary case with SVM as the loss function. $\square$

## 2 Derivation of Implementations

**Derivation of the implementation for linear learning**: Introducing Lagrangian variable $\lambda$, slack variables $\xi_{ij}$ and $\eta_{ij}$, we have that (3) is equivalent to

$$\min_{\boldsymbol{f} \in \mathcal{F}} \frac{n\lambda}{2} \sum_{q=1}^{k-1} \boldsymbol{\beta}_q^T \boldsymbol{\beta} + \sum_{i=1}^{n} \sum_{j \neq y_i} (\xi_{ij} + \eta_{ij})$$

$$subject\,to \ \xi_{ij} \geq 0,$$
$$\eta_{ij} \geq 0,$$
$$\xi_{ij} - \langle \boldsymbol{f}(\boldsymbol{x}_i), \mathcal{Y}_j \rangle - 1 \geq 0,$$
$$\eta_{ij} - (a-1)\langle \boldsymbol{f}(\boldsymbol{x}_i), \mathcal{Y}_j \rangle \geq 0; \ \ i=1,\ldots,n, \ \ j \neq y_i.$$

Now define the corresponding Lagrangian function $\mathcal{L}$ as

$$\mathcal{L} = \frac{n\lambda}{2} \sum_{q=1}^{k-1} \boldsymbol{\beta}_q^T \boldsymbol{\beta} + \sum_{i=1}^{n} \sum_{j \neq y_i} (\xi_{ij} + \eta_{ij}) - \sum_{i=1}^{n} \sum_{j \neq y_i} \tau_{ij} \xi_{ij} - \sum_{i=1}^{n} \sum_{j \neq y_i} \chi_{ij} \eta_{ij}$$
$$- \sum_{i=1}^{n} \sum_{j \neq y_i} \alpha_{ij} \{\xi_{ij} - \langle \boldsymbol{f}(\boldsymbol{x}_i), \mathcal{Y}_j \rangle - 1\} - \sum_{i=1}^{n} \sum_{j \neq y_i} \gamma_{ij} \{\eta_{ij} - (a-1)\langle \boldsymbol{f}(\boldsymbol{x}_i), \mathcal{Y}_j \rangle\},$$

where $\alpha_{ij}$, $\gamma_{ij}$, $\tau_{ij}$, and $\chi_{ij}$; $i=1,\ldots,n, j=1,\ldots,k$ are the Lagrangian multipliers. Define $A_{ij} = I(j \neq y_i)$. Take partial derivative of $\mathcal{L}$ with respect to $\xi_{ij}$, $\eta_{ij}$ and $\boldsymbol{\beta}_q$, and we have

$$\frac{\partial \mathcal{L}}{\partial \xi_{ij}} = A_{ij} - \alpha_{ij} - \tau_{ij} = 0,$$
$$\frac{\partial \mathcal{L}}{\partial \eta_{ij}} = A_{ij} - \gamma_{ij} - \chi_{ij} = 0,$$
$$\frac{\partial \mathcal{L}}{\partial \boldsymbol{\beta}_q} = n\lambda \boldsymbol{\beta}_q + \sum_{i=1}^{n} \sum_{j \neq y_i} \alpha_{ij} \mathcal{Y}_{j,q} \boldsymbol{x}_i + \sum_{i=1}^{n} \sum_{j \neq y_i} \gamma_{ij}(a-1) \mathcal{Y}_{j,q} \boldsymbol{x}_i$$
$$= n\lambda \boldsymbol{\beta}_q + \sum_{i=1}^{n} \sum_{j \neq y_i} \{\alpha_{ij} + (a-1)\gamma_{ij}\} \mathcal{Y}_{j,q} \boldsymbol{x}_i = \boldsymbol{0},$$

where $\mathcal{Y}_{j,q}$ is the $q^{th}$ element of $\mathcal{Y}_j$. Now one can conclude that $0 \leq \alpha_{ij} \leq A_{ij}$, $0 \leq \gamma_{ij} \leq A_{ij}$; $i = 1, \ldots, n$, $j = 1, \ldots, k$, and

$$\boldsymbol{\beta}_q = -\frac{1}{n\lambda} \sum_{i=1}^n \sum_{j \neq y_i} \{\alpha_{ij} + (a-1)\gamma_{ij}\} \mathcal{Y}_{j,q} \boldsymbol{x}_i. \tag{4}$$

Plugging $\boldsymbol{\beta}$ in $\mathcal{L}$, one can verify that

$$\mathcal{L} = -\frac{n\lambda}{2} \sum_{q=1}^{k-1} \boldsymbol{\beta}_q^T \boldsymbol{\beta}_q + \sum_{i=1}^n \sum_{j \neq y_i} \alpha_{ij}.$$

**Derivation of the implementation for kernel learning**: Next, we briefly discuss the case of kernel learning. Let the kernel function be $K(\cdot, \cdot)$, and the corresponding gram matrix be $\boldsymbol{K} = \left(K(\boldsymbol{x}_i, \boldsymbol{x}_{i'})\right)_{i,i'}$. Without loss of generality, assume that the gram matrix $\boldsymbol{K}$ is invertible. If we penalize the intercepts and choose $J(\boldsymbol{f})$ to be the squared norm of $\boldsymbol{f}$ in the RKHS, the optimization problem (3) can be written as (Kimeldorf and Wahba, 1971).

$$\min_{\boldsymbol{f} \in \mathcal{F}} \frac{n\lambda}{2} \sum_{q=1}^{k-1} \boldsymbol{\theta}_q^T \boldsymbol{K} \boldsymbol{\theta}_q + \frac{n\lambda}{2} \sum_{q=1}^{k-1} \theta_{q,0}^2 + \sum_{i=1}^n \sum_{j \neq y_i} (\xi_{ij} + \eta_{ij})$$

$$subject to \ \xi_{ij} \geq 0,$$
$$\eta_{ij} \geq 0,$$
$$\xi_{ij} - \langle \boldsymbol{f}(\boldsymbol{x}_i), \mathcal{Y}_j \rangle - 1 \geq 0,$$
$$\eta_{ij} - (a-1)\langle \boldsymbol{f}(\boldsymbol{x}_i), \mathcal{Y}_j \rangle \geq 0; \quad i = 1, \ldots, n, \ j \neq y_i,$$

where $f_q(\boldsymbol{x}) = \sum_{i=1}^n K(\boldsymbol{x}_i, \boldsymbol{x}) \theta_{q,i} + \theta_{q,0}$; $q = 1, \ldots, k-1$, and $\theta_{q,i}$ is the $i^{th}$ element of $\boldsymbol{\theta}_q$. Now introduce the Lagrangian multipliers $\alpha_{ij}$, $\gamma_{ij}$, $\tau_{ij}$, and $\chi_{ij}$ as in the linear case, take the partial derivatives with respect to $\boldsymbol{\theta}_q$, $\theta_{q,0}$, $\xi_{ij}$ and $\eta_{ij}$ and set to zero, and we have

$$\boldsymbol{\theta}_q = -\frac{1}{n\lambda} \boldsymbol{K}^{-1} [\sum_{i=1}^n \sum_{j \neq y_i} \{\alpha_{ij} + (a-1)\gamma_{ij}\} \mathcal{Y}_{j,q} \boldsymbol{K}_i],$$

$$\theta_{q,0} = -\frac{1}{n\lambda} \sum_{i=1}^n \sum_{j \neq y_i} \{\alpha_{ij} + (a-1)\gamma_{ij}\} \mathcal{Y}_{j,q},$$

where $\boldsymbol{K}_i$ is the $i^{th}$ column of $\boldsymbol{K}$. Therefore, the optimization problem (3) is equivalent

to

$$\min_{\alpha_{ij},\gamma_{ij}} \frac{1}{2n\lambda} \sum_{q=1}^{k-1} [\sum_{i=1}^{n} \sum_{j \neq y_i} \{\alpha_{ij} + (a-1)\gamma_{ij}\} \mathcal{Y}_{j,q} \boldsymbol{K}_i]^T \boldsymbol{K}^{-1} [\sum_{i=1}^{n} \sum_{j \neq y_i} \{\alpha_{ij} + (a-1)\gamma_{ij}\} \mathcal{Y}_{j,q} \boldsymbol{K}_i]$$

$$+ \frac{1}{2n\lambda} \sum_{q=1}^{k-1} [\sum_{i=1}^{n} \sum_{j \neq y_i} \{\alpha_{ij} + (a-1)\gamma_{ij}\} \mathcal{Y}_{j,q}]^2 - \sum_{i=1}^{n} \sum_{j \neq y_i} \alpha_{ij}$$

$subject to \ 0 \leq \alpha_{ij} \leq A_{ij}, \ 0 \leq \gamma_{ij} \leq A_{ij}; \ i = 1, \ldots, n, \ j = 1, \ldots, k.$ (5)

Because $\boldsymbol{K}_i^T \boldsymbol{K}^{-1} \boldsymbol{K}_j = K(\boldsymbol{x}_i, \boldsymbol{x}_j)$, one can verify that (5) can be solved in an analogous manner as (6).

# 3 Extended Numerical Results

| Example 1, Soft with $a=a_1$ | | | Regular | Reject | R&R | Probability Method | |
|---|---|---|---|---|---|---|---|
| | Proportion | | | Error | | Proportion | Error |
| p1 | 50.91 | | 29.36 | 28.38 | 28.38 | 52.26 | 30.17 |
| p2 | 28.35 | size 2: 22.64 | 45.78 | 44.94 | 1.673 | | |
| | | size 3: 5.714 | | | | | |
| p3 | 20.74 | | 69.79 | - | - | 47.74 | - |
| Overall | 100.0 | | 41.92 | 39.41 | 28.17 | 100.0 | 45.64 |
| Example 1, Soft with $a=a_2$ | | | Regular | Reject | R&R | Probability Method | |
| | Proportion | | | Error | | Proportion | Error |
| p1 | 49.43 | | 28.80 | 27.58 | 27.58 | 52.26 | 30.17 |
| p2 | 28.97 | size 2: 24.62 | 45.89 | 45.35 | 1.581 | | |
| | | size 3: 4.349 | | | | | |
| p3 | 21.60 | | 69.61 | - | - | 47.74 | - |
| Overall | 100.0 | | 41.92 | 39.32 | 27.47 | 100.0 | 45.64 |

Table 1: Simulation results for Example 1, the Soft loss.

| Example 1, DWD with $a=a_1$ | | | Regular | Reject | R&R | Probability Method | |
|---|---|---|---|---|---|---|---|
| | Proportion | | | Error | | Proportion | Error |
| p1 | 52.36 | | 31.19 | 29.70 | 29.70 | 58.71 | 31.25 |
| p2 | 27.79 | size 2: 21.60 | 47.07 | 45.37 | 2.011 | | |
| | | size 3: 6.192 | | | | | |
| p3 | 19.85 | | 69.95 | - | - | 41.29 | - |
| Overall | 100.0 | | 42.87 | 39.96 | 28.04 | 100.0 | 43.93 |
| Example 1, DWD with $a=a_2$ | | | Regular | Reject | R&R | Probability Method | |
| | Proportion | | | Error | | Proportion | Error |
| p1 | 49.45 | | 31.79 | 31.30 | 31.30 | 58.71 | 31.25 |
| p2 | 26.62 | size 2: 17.91 | 45.04 | 46.59 | 2.980 | | |
| | | size 3: 8.711 | | | | | |
| p3 | 23.93 | | 66.14 | - | - | 41.29 | - |
| Overall | 100.0 | | 42.87 | 41.80 | 30.60 | 100.0 | 43.93 |

Table 2: Simulation results for Example 1, the DWD loss.

| Example 2, Soft with $a = a_1$ | Regular | Reject | R&R | Probability Method | |
|---|---|---|---|---|---|
| | Proportion | Error | | | Proportion | Error |
| p1 | 57.45 | 27.10 | 26.99 | 26.99 | | |
| p2 | size2: 16.81 ⌊ $\{1,2\}$: 72.2% | 48.34 | 48.35 | 9.141 | 58.61 | 28.74 |
| p3 | 25.74 | 52.72 | - | - | 41.39 | - |
| Overall | 100.0 | 37.94 | 37.43 | 29.92 | 100.0 | 38.33 |
| Example 2, Soft with $a = a_2$ | Regular | Reject | R&R | Probability Method | |
| | Proportion | Error | | | Proportion | Error |
| p1 | 55.21 | 26.58 | 26.37 | 26.37 | | |
| p2 | size2: 16.11 ⌊ $\{1,2\}$: 72.8% | 48.24 | 48.22 | 8.979 | 58.61 | 28.74 |
| p3 | 28.68 | 53.01 | - | - | 41.39 | - |
| Overall | 100.0 | 37.94 | 37.16 | 30.32 | 100.0 | 38.33 |

Table 3: Simulation results for Example 2, the Soft loss.

| Example 2, SVM with $a = a_1$ | Regular | Reject | R&R | Probability Method | |
|---|---|---|---|---|---|
| | Proportion | Error | | | Proportion | Error |
| p1 | 52.53 | 28.15 | 27.81 | 27.81 | | |
| p2 | size2: 12.90 ⌊ $\{1,2\}$: 73.1% | 48.57 | 51.00 | 11.26 | 42.40 | 25.85 |
| p3 | 34.57 | 53.01 | - | - | 57.60 | - |
| Overall | 100.0 | 39.57 | 38.91 | 33.33 | 100.0 | 40.66 |
| Example 2, SVM with $a = a_2$ | Regular | Reject | R&R | Probability Method | |
| | Proportion | Error | | | Proportion | Error |
| p1 | 50.40 | 27.75 | 28.19 | 28.19 | | |
| p2 | size2: 14.41 ⌊ $\{1,2\}$: 73.3% | 47.23 | 50.61 | 11.32 | 42.40 | 25.85 |
| p3 | 35.19 | 53.67 | - | - | 57.60 | - |
| Overall | 100.0 | 39.57 | 38.86 | 33.43 | 100.0 | 40.66 |

Table 4: Simulation results for Example 2, the SVM hinge loss.

| Example 3, DWD with $a = a_1$ | | Regular | Reject | R&R | Probability Method | |
|---|---|---|---|---|---|---|
| | Proportion | | Error | | Proportion | Error |
| p1 | 47.08 | 27.30 | 27.48 | 27.48 | | |
| p2 | 26.25 | size 2: 22.42<br>└ $\{1,2\}$: 42.5%<br>└ $\{3,4\}$: 41.6%<br>size 3: 3.835 | 35.29 | 36.86 | 3.768 | 36.97 | 24.72 |
| p3 | 26.67 | 57.39 | - | - | 63.03 | - |
| Overall | 100.0 | 36.44 | 35.76 | 27.24 | 100.0 | 41.78 |
| Example 3, DWD with $a = a_2$ | | Regular | Reject | R&R | Probability Method | |
| | Proportion | | Error | | Proportion | Error |
| p1 | 45.58 | 25.68 | 26.11 | 26.11 | | |
| p2 | 23.26 | size 2: 19.71<br>└ $\{1,2\}$: 40.3%<br>└ $\{3,4\}$: 42.9%<br>size 3: 3.549 | 36.45 | 35.71 | 1.771 | 36.97 | 24.72 |
| p3 | 31.16 | 56.02 | - | - | 63.03 | - |
| Overall | 100.0 | 36.44 | 35.97 | 27.90 | 100.0 | 41.78 |

Table 5: Simulation results for Example 3, the DWD loss.

| Example 3, SVM with $a = a_1$ | | Regular | Reject | R&R | Probability Method | |
|---|---|---|---|---|---|---|
| | Proportion | | Error | | Proportion | Error |
| p1 | 41.49 | 27.47 | 27.29 | 27.29 | | |
| p2 | 31.26 | size 2: 28.71<br>└ $\{1,2\}$: 41.1%<br>└ $\{3,4\}$: 41.0%<br>size 3: 2.552 | 33.84 | 33.87 | 2.494 | 33.41 | 24.72 |
| p3 | 27.25 | 57.27 | - | - | 66.59 | - |
| Overall | 100.0 | 36.69 | 35.13 | 25.71 | 100.0 | 44.53 |
| Example 3, SVM with $a = a_2$ | | Regular | Reject | R&R | Probability Method | |
| | Proportion | | Error | | Proportion | Error |
| p1 | 42.52 | 26.82 | 26.67 | 26.67 | | |
| p2 | 28.18 | size 2: 24.67<br>└ $\{1,2\}$: 42.2%<br>└ $\{3,4\}$: 41.8%<br>size 3: 3.509 | 35.63 | 36.98 | 3.619 | 33.41 | 24.72 |
| p3 | 29.30 | 56.02 | - | - | 66.59 | - |
| Overall | 100.0 | 36.69 | 35.71 | 26.99 | 100.0 | 44.53 |

Table 6: Simulation results for Example 3, the SVM hinge loss.

| Example 1 | $a = a_1 = 1.333$ | $a = 1.555$ | $a = 1.777$ | $a = a_2 = 2$ |
|---|---|---|---|---|
| Soft | 39.41 | 39.39 | 39.35 | 39.32 |
| DWD | 39.96 | 40.22 | 41.77 | 41.80 |
| Example 2 | $a = a_1 = 1.5$ | $a = 1.667$ | $a = 1.833$ | $a = a_2 = 2$ |
| Soft | 37.43 | 37.39 | 37.26 | 37.16 |
| SVM | 38.91 | 39.01 | 38.81 | 38.86 |
| Example 3 | $a = a_1 = 1.667$ | $a = 1.889$ | $a = 2.111$ | $a = a_2 = 2.333$ |
| DWD | 35.76 | 35.77 | 35.80 | 35.97 |
| SVM | 35.13 | 35.58 | 35.55 | 35.71 |

Table 7: The average empirical 0-$d$-1 loss on the test data sets for simulated Examples 1-3 using different loss functions and various $a$ values.

| GBM, Soft with $a = a_1$ | | Regular | Reject | R&R | Probability Method | |
|---|---|---|---|---|---|---|
| | Proportion | | Error | | Proportion | Error |
| p1 | 72.15 | 13.69 | 13.69 | 13.69 | | |
| p2 | 18.99 | size 2: 17.14 ⌊ $\{C,M\}$: 44.3% ⌊ $\{N,P\}$: 33.7% size 3: 1.853 | 41.35 | 39.53 | 3.724 | 72.58 | 13.78 |
| p3 | 8.857 | 43.33 | - | - | 27.42 | - |
| Overall | 100.0 | 21.85 | 20.97 | 14.13 | 100.0 | 21.25 |
| GBM, Soft with $a = a_2$ | | Regular | Reject | R&R | Probability Method | |
| | Proportion | | Error | | Proportion | Error |
| p1 | 70.15 | 13.16 | 12.84 | 12.84 | | |
| p2 | 19.85 | size 2: 18.42 ⌊ $\{C,M\}$: 44.8% ⌊ $\{N,P\}$: 35.8% size 3: 1.428 | 40.09 | 42.35 | 4.055 | 72.58 | 13.78 |
| p3 | 10.00 | 54.86 | - | - | 27.42 | - |
| Overall | 100.0 | 21.85 | 21.00 | 13.81 | 100.0 | 21.25 |

Table 8: Data analysis results for the GBM data, the Soft loss.

| ZIP, DWD with $a = a_1$ | | Regular | Reject | R&R | Probability Method | |
|---|---|---|---|---|---|---|
| | Proportion | | Error | | Proportion | Error |
| p1 | 97.05 | 2.087 | 2.087 | 2.087 | | |
| p2 | size 2: 2.578 ⌊ $\{4,9\}$: 55.9% | 35.71 | 28.57 | 0.110 | 98.90 | 2.607 |
| p3 | 0.368 | 95.14 | - | - | 1.104 | - |
| Overall | 100.0 | 3.314 | 2.909 | 2.175 | 100.0 | 3.020 |
| ZIP, DWD with $a = a_2$ | | Regular | Reject | R&R | Probability Method | |
| | Proportion | | Error | | Proportion | Error |
| p1 | 97.47 | 2.277 | 2.166 | 2.166 | | |
| p2 | size 2: 2.119 ⌊ $\{4,9\}$: 57.2% | 28.57 | 21.42 | 0.150 | 98.90 | 2.607 |
| p3 | 0.412 | 97.25 | - | - | 1.104 | - |
| Overall | 100.0 | 3.314 | 2.885 | 2.279 | 100.0 | 3.020 |

Table 9: Data analysis results for the ZIP data, the DWD loss.

# References

Bartlett, P. L. and Wegkamp, M. H. (2008), "Classification with a Reject Option Using a Hinge Loss," *Journal of Machine Learning Research*, 9, 1823–1840.

Kimeldorf, G. and Wahba, G. (1971), "Some Results on Tchebycheffian Spline Functions," *Journal of Mathematical Analysis and Applications*, 33, 82–95.

van der Vaart, A. W. and Wellner, J. A. (2000), *Weak Convergence and Empirical Processes with Application to Statistics*, Springer, 1st ed.

Wang, L. and Shen, X. (2007), "On $L_1$-norm Multi-class Support Vector Machines: Methodology and Theory," *Journal of the American Statistical Association*, 102, 595–602.

Zhang, C. and Liu, Y. (2014), "Multicategory Angle-based Large-margin Classification," *Biometrika*, 101, 625–640.

Zhang, C., Liu, Y., and Wu, Y. (2016), "On Quantile Regression in Reproducing Kernel Hilbert Spaces with Data Sparsity Constraint," *Journal of Machine Learning Research*, 17, 1–45.

Zhou, D.-X. (2002), "The Covering Number in Learning Theory," *Journal of Complexity*, 18, 739–767.



\end{document}